\documentclass[
    10pt,
    twocolumn,
    letterpaper,
    ]{article}

\usepackage[pagenumbers]{cvpr}              

\usepackage[dvipsnames]{xcolor}
\usepackage{wrapfig}
\usepackage{ulem}

\usepackage{pgfplots}
\usepgfplotslibrary[groupplots,dateplot]
\usetikzlibrary[patterns,shapes.arrows]
\usetikzlibrary{matrix}
\pgfplotsset{compat=newest}

\usepackage{xfrac}
\usepackage{multirow}
\usepackage{makecell}

\usepackage{subcaption}
\usepackage[export]{adjustbox}

\usepackage[accsupp]{axessibility}

\newlength{\figh}
\newlength{\figw}
\newlength{\figwh}
\newlength{\figww}
\newcommand{\nside}{n_{\mathrm{side}}}
\newcommand{\npix}{n_{\mathrm{pix}}}
\newcommand{\npatch}{n_{\mathrm{patch}}}
\newcommand{\nwindow}{n_{\mathrm{win}}}
\newcommand{\nshift}{n_{\mathrm{shift}}}

\renewcommand{\emph}[1]{\textit{#1}}

\definecolor{cvprblue}{rgb}{0.21,0.49,0.74}
\usepackage[
    pagebackref,
    breaklinks,
    colorlinks,
    citecolor=cvprblue
    ]{hyperref}

\def\paperID{15658} 
\def\confName{CVPR}
\def\confYear{2024}

\title{HEAL-SWIN: A Vision Transformer On The Sphere}

\makeatletter
\def\@fnsymbol#1{\ensuremath{\ifcase#1\or *\or a\or
   b\or c\or d\or e\or \dagger\dagger
   \or \ddagger\ddagger \else\@ctrerr\fi}}
\makeatother
    
\author{
  Oscar Carlsson\thanks{Equal contribution}\ \,\thanks{Department of Mathematical Sciences, Chalmers University of Technology, University of Gothenburg, SE-412 96 Gothenburg, Sweden}\ \,\thanks{Corresponding author, email: \href{mailto:osccarls@chalmers.se}{osccarls@chalmers.se}} \\
  \and
  Jan E.\ Gerken\footnotemark[1]{\ \,}\footnotemark[2]\\
  \and
  Hampus Linander\,\footnotemark[2]\\
  \and
  Heiner Spieß\,\thanks{Neural Information Processing, Science of Intelligence, Technical University Berlin, DE-10623 Berlin, Germany}\\
  \and
  Fredrik Ohlsson\,\thanks{Department of Mathematics and Mathematical Statistics, Umeå University, SE-901 87 Umeå, Sweden}\\
  \and
  Christoffer Petersson\thanks{Zenseact, SE-417 56 Gothenburg, Sweden}{\ \,}\footnotemark[2]\\
  \and
  Daniel Persson\footnotemark[2] \\
}

\begin{document}
\maketitle

\begin{abstract}
High-resolution wide-angle fisheye images are becoming more and more important for robotics applications such as autonomous driving. However, using ordinary convolutional neural networks or vision transformers on this data is problematic due to projection and distortion losses introduced when projecting to a rectangular grid on the plane. We introduce the HEAL-SWIN transformer, which combines the highly uniform Hierarchical Equal Area iso-Latitude Pixelation (HEALPix) grid used in astrophysics and cosmology with the  Hierarchical Shifted-Window (SWIN) transformer to yield an efficient and flexible model capable of training on high-resolution, distortion-free spherical data. In HEAL-SWIN, the nested structure of the HEALPix grid is used to perform the patching and windowing operations of the SWIN transformer, enabling the network to process spherical representations with minimal computational overhead. We demonstrate the superior performance of our model on both synthetic and real automotive datasets, as well as a selection of other image datasets, for semantic segmentation, depth regression and classification tasks. Our code is publicly available\footnote{https://github.com/JanEGerken/HEAL-SWIN}. 
\end{abstract}

\section{Introduction}

\begin{figure}
    
    \centering
    
    \includegraphics[width=0.4\textwidth]{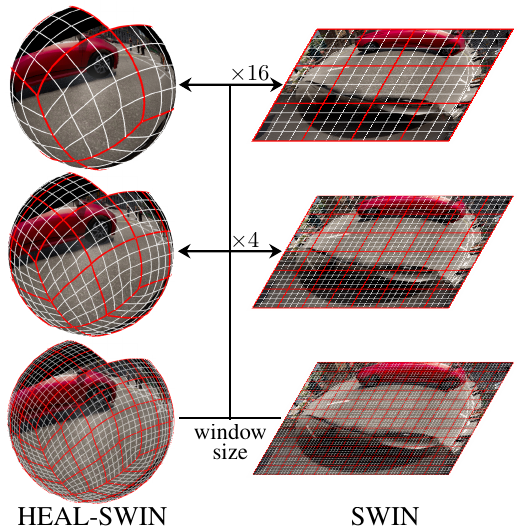}
    
    \caption{Our HEAL-SWIN model uses the nested structure of the HEALPix grid to lift the windowed self-attention of the SWIN model onto the sphere.}
    
    \label{fig:HEAL-SWIN}
    
\end{figure}
High-resolution fisheye cameras are among the most common and important sensors in modern intelligent vehicles~\cite{9913937}. Due to their non-rectilinear mapping functions and large field of view, fisheye images are highly distorted. Moreover, the most commonly used large-scale computer vision and autonomous driving datasets do not contain fisheye images. For these reasons, fisheye images have received much less attention than rectilinear images in the literature.

Despite the distortions introduced by the mapping function, the traditional approach for dealing with this kind of data is to use standard (flat) convolutional neural networks which are adjusted to the distortions and either preprocess the data~\cite{zhang2019b, leeSpherePHDApplyingCNNs2019,haim2019surface,eder2019convolutions,duSphericalTransformerAdapting2021,pmlr-v139-shakerinava21a} or deform the convolution kernels~\cite{tatenoDistortionAwareConvolutionalFilters2018}. However, these approaches struggle to capture the inherent spherical geometry of the images since they operate on a flat approximation of the sphere. Errors and artifacts arising from handling the strong and spatially inhomogeneous distortions are particularly problematic in safety-critical applications such as autonomous driving.

Utilizing spherical representations is an approach taken by some models~\cite{cohen2018b, esteves2020c, cobb2021} which lift convolutions to the sphere. These models rely on a rectangular grid in spherical coordinates, namely the Driscoll--Healy grid~\cite{driscoll1994}, to perform efficient Fourier transforms. However, this approach has several disadvantages for high-resolution data: First, the sampling in this grid is not uniform but much denser at the poles, necessitating very high bandwidths to resolve fine details around the equator. Second, the Fourier transforms in the aforementioned models require tensors in the Fourier domain of the rotation group $\mathrm{SO}(3)$ which scale with the third power of the bandwidth, limiting the resolution. Third, the Fourier transform is very tightly coupled to the input domain: If the input data lies only on a half-sphere, as is often the case for fisheye images, the definition of the convolutional layers needs to be changed to use this data efficiently.

As a novel way to address all these problems at the same time, we propose to combine an adapted vision transformer with the Hierarchical Equal Area iso-Latitude Pixelisation (HEALPix) grid~\cite{gorski1998analysis}. The HEALPix grid was developed for capturing the high-resolution measurements of the cosmic microwave background performed by the MAP and PLANCK satellites featuring a uniform distribution of grid points on the sphere that associates the same area to each pixel. This is in contrast to most other grids used in the literature like the Driscoll--Healy grid or the icosahedral grid.

In our model, which we call HEAL-SWIN, we use a modified version of the Hierarchical Shifted-Window (SWIN) transformer~\cite{liu2021Swin} to learn directly on the HEALPix grid with minimal computational overhead. The SWIN transformer performs attention over blocks of pixels called \emph{windows} which aligns well with the nested structure of the HEALPix grid (Figure~\ref{fig:HEAL-SWIN}). To distribute information globally, the SWIN transformer shifts the windows in every other layer, creating overlapping regions. In HEAL-SWIN, we employ the same principles but tailor them to fit the structure of the HEALPix grid. In particular, we propose two different strategies for shifting windows on the sphere: Either aligned with the hierarchical structure of the HEALPix grid or in a spiral from one pole to the other.

Besides excellent performance, an additional benefit of using HEAL-SWIN is that the attention layers do not require a Fourier transform and can therefore easily deal with high-resolution data and with data which covers only part of the sphere, resulting in significant efficiency gains. This is of central importance to dealing with high-resolution fisheye images which cover about half of the sphere, leaving half of the input unused in spherical models which rely on operating on the entire sphere.

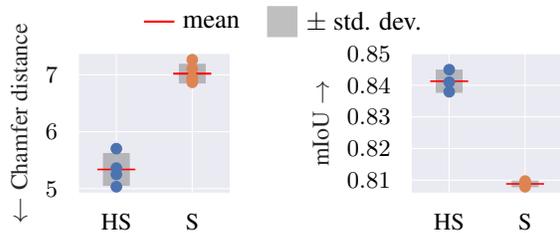
\begin{figure}[t]
    \centering
    \begin{subfigure}[b]{0.9\columnwidth}
        \centering
        % This file was created with tikzplotlib v0.10.1.

\definecolor{darkslategray38}{RGB}{38,38,38}
\definecolor{gray}{RGB}{128,128,128}
\definecolor{lavender234234242}{RGB}{234,234,242}
\definecolor{steelblue76114176}{RGB}{76,114,176}

\begin{tikzpicture}[
  blacknode/.style={shape=rectangle, draw=gray, fill=gray, opacity=0.5},
  rednode/.style={shape=circle, draw=red, line width=2}
]
\matrix[ampersand replacement=\&, ] at (0.0,0.0){
\node[] (l) {\phantom{a}}; \& \node[] {mean \phantom{a}}; \&
\node[shape=rectangle, draw=gray, fill=gray, opacity=0.5] {\phantom{a}};\& \node[] {$\pm$ std. dev.}; \\
};
\draw[red, thick] (l.west)--(l.east);
\end{tikzpicture}
        \vspace{-5mm}
    \end{subfigure}
    
    \begin{subfigure}{0.45\columnwidth}
        \definecolor{darkslategray38}{RGB}{38,38,38}
\definecolor{gray}{RGB}{128,128,128}
\definecolor{lavender234234242}{RGB}{234,234,242}
\definecolor{steelblue76114176}{RGB}{76,114,176}

\begin{tikzpicture}[
  blacknode/.style={shape=rectangle, draw=gray, fill=gray, opacity=0.5},
  rednode/.style={shape=circle, draw=red, line width=2}
]
\pgfplotsset{every tick label/.append style={font=\small}}

\begin{axis}[
axis background/.style={fill=lavender234234242},
axis line style={white},
height=0.8\figh,
tick align=outside,
width=0.6\figw,
x grid style={white},
xmajorgrids,
xmajorticks=true,
xtick style={draw=none},
xmin=-1, xmax=1,
xtick style={color=darkslategray38},
xtick={-0.5,0.5},
xticklabels={HS,S},
y grid style={white},
ylabel=\textcolor{darkslategray38}{$\leftarrow$ \small Chamfer distance},
ymajorgrids,
ymajorticks=true,
ytick style={draw=none},
ymin=4.9214, ymax=7.3766, 
ytick style={color=darkslategray38}
]
\draw[draw=gray,fill=gray,opacity=0.5] (axis cs:-0.67,5.05734353252462) rectangle (axis cs:-0.33,5.61565646747538);
\draw[draw=gray,fill=gray,opacity=0.5] (axis cs:0.33,6.85640518300873) rectangle (axis cs:0.67,7.18439481699127);
\addplot [
  mark=*,
  only marks,
  scatter,
  scatter/@post marker code/.code={
  \endscope
},
  scatter/@pre marker code/.code={
  \expanded{
  \noexpand\definecolor{thispointdrawcolor}{RGB}{\drawcolor}
  \noexpand\definecolor{thispointfillcolor}{RGB}{\fillcolor}
  }
  \scope[draw=thispointdrawcolor, fill=thispointfillcolor]
},
  visualization depends on={value \thisrow{draw} \as \drawcolor},
  visualization depends on={value \thisrow{fill} \as \fillcolor}
]
table{
x  y  draw  fill
-0.5 5.25 76,114,176 76,114,176
-0.5 5.033 76,114,176 76,114,176
-0.5 5.361 76,114,176 76,114,176
-0.5 5.702 76,114,176 76,114,176
0.5 6.949 221,132,82 221,132,82
0.5 6.868 221,132,82 221,132,82
0.5 7.108 221,132,82 221,132,82
0.5 6.912 221,132,82 221,132,82
0.5 7.265 221,132,82 221,132,82
};
\addplot [semithick, red]
table {
-0.75 5.3365
-0.25 5.3365
};
\addplot [thick, red]
table {
0.25 7.0204
0.75 7.0204
};

\end{axis}
\end{tikzpicture}
    \end{subfigure}
    ~
    \begin{subfigure}{0.45\columnwidth}
        \begin{tikzpicture}

\definecolor{darkslategray38}{RGB}{38,38,38}
\definecolor{gray}{RGB}{128,128,128}
\definecolor{lavender234234242}{RGB}{234,234,242}

\pgfplotsset{every tick label/.append style={font=\small}}
\begin{axis}[
axis background/.style={fill=lavender234234242},
axis line style={white},
height=0.8\figh,
tick align=outside,
width=0.6\figw,
x grid style={white},
xmajorgrids,
xmajorticks=true,
xtick style={draw=none},
xmin=-1, xmax=1,
xtick style={color=darkslategray38},
xtick={-0.5,0.5},
xticklabels={HS,S},
y grid style={white},
ylabel=\textcolor{darkslategray38}{\small mIoU $\rightarrow$},
ymajorgrids,
ymajorticks=true,
ytick style={draw=none},
ymin=0.8059260815382, ymax=0.850047531723976,
ytick style={color=darkslategray38}
]
\draw[draw=gray,fill=gray,opacity=0.5] (axis cs:-0.67,0.837761485533029) rectangle (axis cs:-0.33,0.844763275031139);
\draw[draw=gray,fill=gray,opacity=0.5] (axis cs:0.33,0.807939443032147) rectangle (axis cs:0.67,0.809722788095274);
\addplot [
  mark=*,
  only marks,
  scatter,
  scatter/@post marker code/.code={
  \endscope
},
  scatter/@pre marker code/.code={
  \expanded{
  \noexpand\definecolor{thispointdrawcolor}{RGB}{\drawcolor}
  \noexpand\definecolor{thispointfillcolor}{RGB}{\fillcolor}
  }
  \scope[draw=thispointdrawcolor, fill=thispointfillcolor]
},
  visualization depends on={value \thisrow{draw} \as \drawcolor},
  visualization depends on={value \thisrow{fill} \as \fillcolor}
]
table{
x  y  draw  fill
-0.5 0.837924659252167 76,114,176 76,114,176
-0.5 0.844906330108643 76,114,176 76,114,176
-0.5 0.840956151485443 76,114,176 76,114,176
};
\addplot [
  mark=*,
  only marks,
  scatter,
  scatter/@post marker code/.code={
  \endscope
},
  scatter/@pre marker code/.code={
  \expanded{
  \noexpand\definecolor{thispointdrawcolor}{RGB}{\drawcolor}
  \noexpand\definecolor{thispointfillcolor}{RGB}{\fillcolor}
  }
  \scope[draw=thispointdrawcolor, fill=thispointfillcolor]
},
  visualization depends on={value \thisrow{draw} \as \drawcolor},
  visualization depends on={value \thisrow{fill} \as \fillcolor}
]
table{
x  y  draw  fill
0.5 0.80793160200119 221,132,82 221,132,82
0.5 0.808847010135651 221,132,82 221,132,82
0.5 0.809714734554291 221,132,82 221,132,82
};
\addplot [semithick, red]
table {
-0.75 0.841262380282084
-0.25 0.841262380282084
};
\addplot [semithick, red]
table {
0.25 0.808831115563711
0.75 0.808831115563711
};

\end{axis}

\end{tikzpicture}
    \end{subfigure}
    \caption{Chamfer distance (lower is better) and mIoU (higher is better) for HEAL-SWIN (HS) and SWIN (S). 
    Details are provided in Section~\ref{sec:depth-estimation} and in Section~\ref{sec:seg-synthetic-images}.}
    
    \label{fig:main_results}
\end{figure}

In order to verify the efficacy of our proposed model, we train HEAL-SWIN on a number of different computer vision tasks: classification, depth estimation and semantic segmentation of fisheye images from a diverse range of datasets. In these experiments, we put particular emphasis on tasks from the autonomous driving domain for which high-resolution input images and very precise outputs in terms of 3D information are critical, necessitating geometry-aware representations.
To the best of our knowledge, we are the first to treat fisheye images in automotive applications as spherical signals.

In particular, we perform fisheye-image-based depth estimation and 
compare the resulting predicted 3D point clouds to the corresponding ground truth point clouds. In this way, we can assess the quality of the learned representations for downstream tasks that rely on accurate 3D information, such as general obstacle detection and collision avoidance, allowing us to isolate the effect of using spherical representations on the HEALPix grid.
We show that our model outperforms the flat SWIN transformer in this setting on the SynWoodScape dataset~\cite{sekkat2022} of computer-generated fisheye images of street scenes, see Figure~\ref{fig:main_results} (left), 
establishing the benefit of using HEALPix representations. For the semantic segmentation task, we confirm the superior performance of our model on both the real-world WoodScape dataset~\cite{woodscape2019} and on the SynWoodScape dataset, see Figure~\ref{fig:main_results} (right)
. For both depth estimation and semantic segmentation we optimize the loss directly on the sphere. 

Having shown that the addition of spherical representations is beneficial for high-resolution image datasets, we verify that our model is competitive with the state of the art in spherical image classification and semantic segmentation of indoor scenes.

Although we focus in this work on fisheye images, our model is not specific to this kind of problem but can be trained on any high-resolution spherical data, as provided for instance by satellites mapping the sky or the earth. In particular, deep spherical models (and in particular transformers) were recently shown to outperform physical models for weather prediction tasks~\cite{lam2022, bi2023, kurth2023, nguyen2023a}, opening another important application domain for our model.

Our main contributions are as follows:
\begin{itemize}

\item 
We construct the HEAL-SWIN transformer which operates on high-resolution spherical representations by combining the spherical HEALPix grid with an adapted SWIN transformer. By exploiting the similar hierarchical structures in HEALPix and SWIN, we construct windowing and shifting mechanisms for the HEALPix grid which efficiently deal with data that only covers part of the sphere.

\item 
We treat fisheye images in automotive applications for the first time as distortion-free spherical signals. We demonstrate the superiority of this approach for depth estimation and semantic segmentation on both synthetic and real automotive datasets.
\item To compare HEAL-SWIN to other models operating on spherical representations, we benchmark on the Stanford 2D-3D-S indoor fisheye dataset ~\cite{Armeni2017Stanford} and find that our model outperforms comparable spherical models. 

\end{itemize}

\section{Related work}
The transformer architecture was introduced in~\cite{Vaswani2017Attention}, extended to the Vision Transformer (ViT) in~\cite{Dosovitskiy2021VisionTransformer}, and further refined in~\cite{liu2021Swin} to the Shifted-Window (SWIN) transformer based on attention in local windows combined with window shifting to account for global structure. A first step towards spherical transformer models has been proposed in~\cite{Cho2022SphericalTransformer} where various spherical grids are used to extract patches on which the ViT is applied, and in~\cite{Ruhling2022ClimFormer} where icosahedral grid sampling is combined with the Adaptive Fourier Neural Operator (AFNO, see~\cite{Guibas2022AdaptiveFourierNeuralOperators}) architecture for spatial token mixing to account for the geometry of the sphere. Compared to previous spherical transformer designs, such as \cite{Cho2022SphericalTransformer}, our model constitutes a significant improvement by incorporating local attention and window shifting to accommodate high-resolution images in the spherical geometry, without requiring the careful construction of accurate graph representations of the spherical geometry. 

The SWIN paradigm has been incorporated into transformers operating on 3D point clouds (e.g.~LiDAR depth data) by combining voxel based models (see~\cite{Mao2021VoxelTransformer}) with sparse~\cite{Fan2022SingelStrideSparseTransformer} and stratified~\cite{Lai2022StratifiedTransformer} local attention mechanisms. In~\cite{Lai2023SphericalTransformer3DRecognition}, spherical geometry is used to account for long range interactions in the point cloud to create a transformer based on local self-attention in radial windows. Closer in spirit to our approach is~\cite{Guo2023SphericalWindowBasedPCTransformer}, where the point cloud is projected to the sphere, and partitioned into neighborhoods. Local self-attention and patch merging is then applied to the corresponding subsets of the point cloud, and shifting of subsets is achieved by rotations of the underlying sphere. Our HEAL-SWIN model inherits windowing and shifting from the SWIN architecture, but in contrast to the point cloud based approaches we handle data native to the sphere and use the HEALPix grid to construct a hierarchical sampling scheme which minimizes distortions, allows for the handling of high-resolution data, and, in addition, incorporates new shifting strategies specifically adapted to the HEALPix grid.

Several Convolutional Neural Network (CNN) based models have been proposed to accommodate spherical data. As mentioned above,~\cite{cohen2018b, esteves2020c, cobb2021} use an equirectangular grid and implement convolutions in Fourier space, while \cite{Krachmalnicoff2019} applies CNNs directly to the HEALPix partitions of the sphere. DeepSphere is a graph-based CNN which is rotationally equivariant for radial filters~\cite{defferrard2020}. The underlying graph allows for a non-uniform sampling which can be beneficial for certain applications. DeepSphere has been specialized to the HEALPix grid~\cite{perraudin2019} but cannot exploit the nested grid in the same way as HEAL-SWIN which employs windowed self-attention.

Other works that consider spherical CNNs combined with the HEALPix grid include  \cite{Jiang20219SphericalCNNUnstructuredGrids,duSphericalTransformerAdapting2021}. 
Compared to previous works combining CNNs and the HEALPix grid, the windowed self-attention equips our model with the ability to efficiently encode long-range interactions. Moreover, transformers are known to be able to benefit from large datasets where CNNs reach their capacity limits.

Due to the general shortage of computer vision datasets involving fisheye images, it is common to evaluate proposed methods by creating spherical versions of MNIST \cite{6296535} and SYNTHIA \cite{7780721}. In 2021, the first set of computer vision tasks for real-world automotive fisheye images was released for the WoodScape dataset~\cite{woodscape2019}. In 2022, the SynWoodScape~\cite{sekkat2022} dataset was released, consisting of synthetic fisheye images generated using the driving simulator CARLA~\cite{carla17}.

Outside of the automotive domain, a common benchmark for fishey image data is the Stanford
2D-3D dataset~\cite{Armeni2017Stanford} of indoor scenes. Previous works that, like our proposed HEAL-SWIN model, perform evaluation in the spherical domain for this dataset include;~\cite{zhang2019b} based on an icosahedral grid sampling,~\cite{Hart2023Interpolated} which constructs structured graph convolutions to incorporate the spherical geometry,~\cite{Jiang20219SphericalCNNUnstructuredGrids} where convolutional kernels are parameterised using differential operators on the sphere, and~\cite{esteves2020c} which uses spin-weighted spherical harmonics to perform spherical convolutions in the Fourier domain.

\section{HEAL-SWIN}
We propose to combine the SWIN-transformer~\cite{liu2021Swin, liu2021swinv2} with the HEALPix grid~\cite{gorski1998analysis} resulting in the HEAL-SWIN-transformer which is capable of training on high-resolution images on the sphere. In this section, we describe the structure of the HEAL-SWIN model in detail.

\subsection{The SWIN transformer}
\label{sec:swin}

The SWIN-transformer is a computationally efficient vision transformer which attends to windows that are shifted from layer to layer, enabling a global distribution of information while mitigating the quadratic scaling of attention in the number of pixels.

In the first layer of the SWIN-transformer, squares of pixels are joined into tokens called \emph{patches} to reduce the initial resolution of the input images. Each following SWIN-layer consists of two transformer blocks which perform attention over squares of patches called \emph{windows}. The windows are shifted along the patch-grid axes by half a window size, before the attention for the second transformer block is computed. In this way, information is distributed across window boundaries. To down-scale the spatial resolution, two-by-two blocks of patches are periodically merged.

An important detail in this setup is that at the boundary of the image, shifting creates partially-filled windows. Here, the SWIN-transformer fills up the windows with patches from other partially-filled windows and then performs a masked version of self-attention which does not attend to pixel pairs which originated from different regions of the original.

For the depth-estimation and segmentation tasks we consider in this work, we use a UNet-like variant of the SWIN-transformer~\cite{swinunet, bi2023} which extends the encoding layers of the original SWIN-transformer by corresponding decoding layers connected via skip connections. The decoding layers are identical to the encoding layers, only the patch merging layers are replaced by patch expansion layers which expand one patch into a two-by-two block of patches such that the output of the entire model has the same resolution as the input.

\begin{figure}
  \centering
  \includegraphics[width=0.4\linewidth]{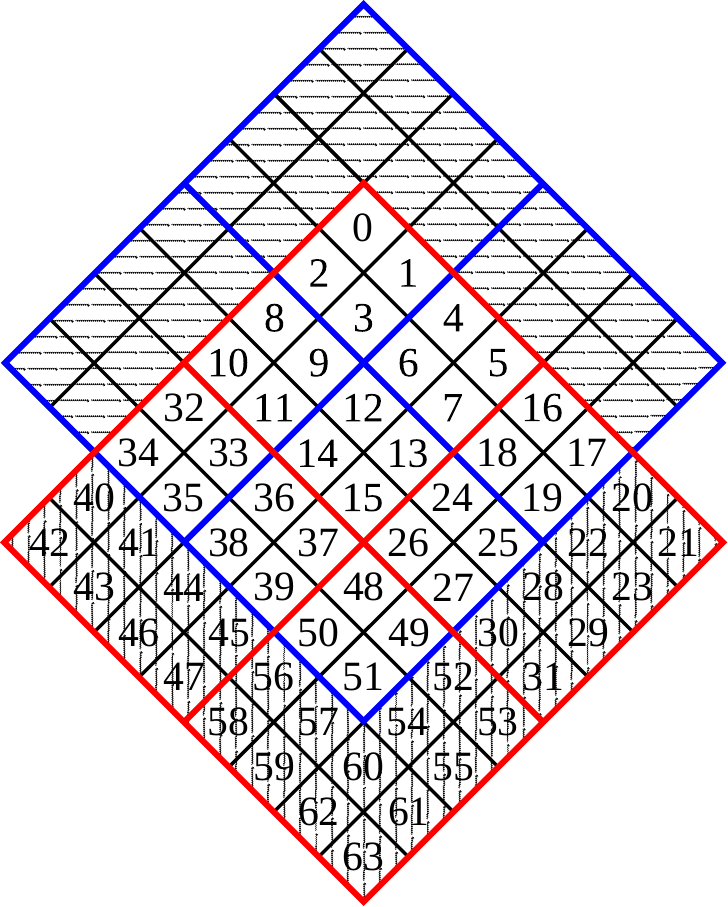}
  \caption{Grid shifting scheme for window size 16: The windows before the shift are framed in red and the patches are numbered in the nested scheme. After a shift by half a window size, the patches are divided into the windows framed in blue, so that e.g. patch 0 becomes patch 12 after the shift. The hashed regions are masked in the attention layer. The patches hashed horizontally correspond to the pixels marked in yellow in Figure~\ref{fig:shifting} (left). They are filled with patches hashed vertically which correspond to the pixels lost in the center of Figure~\ref{fig:shifting} (left).}
  \label{fig:grid_shifting_scheme}
\end{figure}

\subsection{The HEALPix grid}
\label{sec:healpix}
In the HEAL-SWIN-transformer, the patches are not associated to an underlying rectangular pixel grid as in the original SWIN-transformer, but to the HEALPix grid on the sphere. The HEALPix grid is constructed from twelve equal-area, four-sided polygons (quadrilaterals) of different shapes which tessellate the sphere (drawn in red in the top-left sphere of Figure~\ref{fig:HEAL-SWIN}). These are subdivided along their edges $\nside$ times to yield a high-resolution partition of the sphere into $\npix=12\cdot \nside^{2}$
equal-area, iso-latitude quadrilaterals (the $\nside=2$ grid is drawn in white in the top-left sphere of Figure~\ref{fig:HEAL-SWIN}). To allow for a nested (hierarchical) grid structure, $\nside$ needs to be a power of two. The pixels of the grid are then placed at the centers of the quadrilaterals. The resulting positions are sorted in a list either in the \emph{nested} ordering descending from the iterated subdivisions of the base-resolution quadrilaterals, as illustrated in Figure~\ref{fig:grid_shifting_scheme}, or in a \emph{ring} ordering which follows rings of equal latitude from one pole to the other. Given this data structure, we use a one-dimensional version of the SWIN transformer which operates on these lists. For retrieving the positions of the HEALPix pixels at a certain resolution, translating between the nested- and ring indexing and interpolating in the HEALPix grid, we use the Python package \texttt{healpy}~\cite{Zonca2019}.

Since for our experiments, we consider images taken by fisheye cameras which cover only half of the sphere, we use a modification of the HEALPix grid, where we only use the pixels in eight out of the twelve base-resolution quadrilaterals which we will call \emph{base pixels}. These cover approximately half of the sphere and allow for an efficient handling of the input data, in contrast to many methods used in the literature which require a grid covering the entire sphere. The restriction to the first eight base pixels is performed by selecting the first $\sfrac{8}{12}$ entries in the HEALPix grid list in nested ordering.

\subsection{HEAL-SWIN}
In HEAL-SWIN we adapt the patching-, windowing- and shifting mechanisms of the SWIN transformer to the HEALPix grid, enabling the transformer to operate on an inherently spherical representation of the data.

\subsubsection{Patches and windows}
The nested structure of the HEALPix grid aligns very well with the patching, windowing, patch-merging and patch-expansion operations of the SWIN transformer. Correspondingly, the modifications to the SWIN transformer result in a minimal computational overhead.

The input data is provided as a list in the nested ordering described in Section~\ref{sec:healpix} above. Therefore, we start from a one-dimensional version of the model (in contrast to the usual two-dimensional version used for images). Then, the patching of the input pixels amounts to joining $\npatch$ consecutive pixels into a patch, where $\npatch$ is a power of four. Due to the nested ordering and the homogeneity of the HEALPix grid, the resulting patches cover quadrilateral areas of the same size on the sphere. Similarly, to partition patches into windows over which attention is performed, $\nwindow$ consecutive patches are joined together, where $\nwindow$ is again a power of four. E.g.\ in Figure~\ref{fig:grid_shifting_scheme}, a window size of $\nwindow=16$ is illustrated with patches $0-15$ in the first window, patches $16-31$ in the second window etc. For patch merging, we can similarly merge $n=4^{k}$ consecutive patches in the HEALPix list for downscaling and expand $n=4^{k}$ patches for upscaling.

\subsubsection{Shifting} 
\label{sec:shifting}
\begin{figure}
  \centering
  \begin{minipage}{0.48\linewidth}
      \centering
      \includegraphics[width=\linewidth]{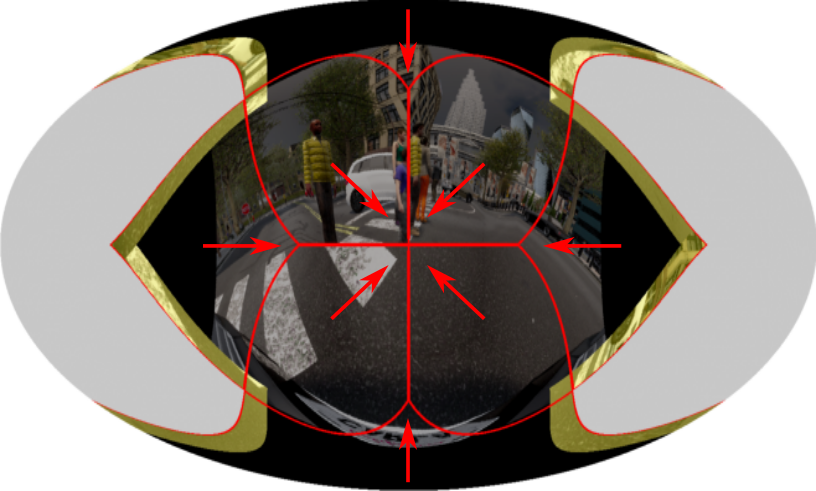}
  \end{minipage}
  \begin{minipage}{0.48\linewidth}
      \centering
      \includegraphics[width=\linewidth]{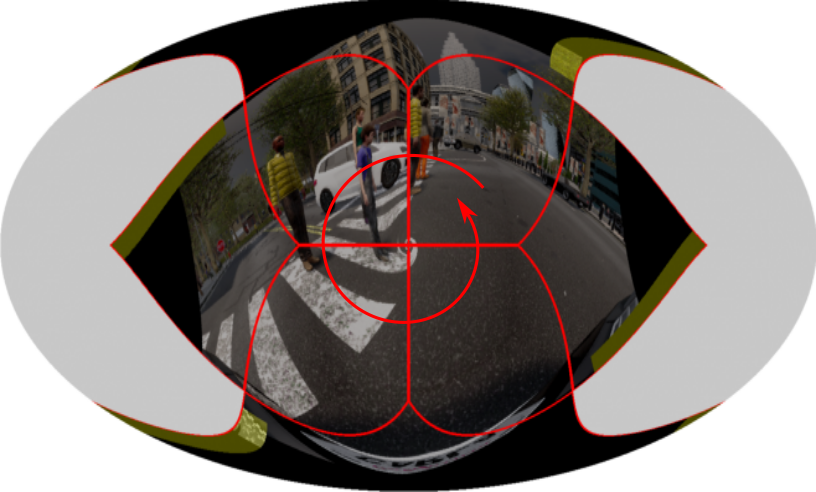}
  \end{minipage}
  \caption{Grid (left) and spiral (right) shifting strategies for the HEAL-SWIN transformer, projected onto the plane for visualization. The grid lines outline the eight base pixels used for this dataset, arrows indicate the directions in which pixels move. Highlighted regions are masked in the attention layers. Note that in grid shifting, pixels at boundaries of colliding base pixels are moved to the outer edge. In ring shifting, distortions are introduced towards the pole (center). For better visibility, the amount of shifting in these images is exaggerated.}
  \label{fig:shifting}
  
\end{figure}

As mentioned above, to distribute information globally in the image, the SWIN transformer shifts the windows by half a window size along both image axes in every second attention layer. We have experimented with two different ways of performing this shifting in the HEALPix grid.

The most direct generalization of the shifting in the pixel grid of the original SWIN-transformer is a shifting in the HEALPix grid along the axes of the quadrilaterals of the base-resolution pixels, cf.\ Figure~\ref{fig:grid_shifting_scheme} and Figure~\ref{fig:shifting} (left). We call this \emph{grid} shifting and shift by half a window in both directions. Similarly to the original SWIN shifting scheme, there are boundary effects at the edge of the half sphere covered by the grid. Additionally, due to the alignment of the base pixels relative to each other, the shifting necessarily clashes at some base-pixel boundaries in the interior of the image. As in the original SWIN transformer, both of these effects are handled by reshuffling the problematic pixels to fill up all windows and subsequently masking the attention mechanism to not attend to pixel pairs which originate from different regions of the sphere. The corresponding pixels are highlighted in yellow in Figure~\ref{fig:shifting}.

In the \emph{spiral} shifting scheme, we first convert the nested ordering into a ring ordering and then perform a roll operation by $\nshift$ on that list. Finally, we convert back to the nested ordering. In this way, windows are shifted along the azimuthal angle by $\nshift$ pixels, with slight distortions which grow larger towards the poles due to the decreased length of circles of constant latitude, cf.\ Figure~\ref{fig:shifting} (right). A shift by half a window is achieved with a shift size of
  $\nshift={\sqrt{\nwindow}}/{2}.$\footnote{Since $\nwindow$ is a power of four, the square root is always an integer.}

As in the grid shifting scheme, we encounter boundary effects. In the spiral shifting, they occur at the pole and at the boundary of the half sphere covered by the grid. These effects are again handled by reshuffling the pixels and masking the attention mechanism appropriately. In this scheme, there are no boundary effects in the interior of the image.

Both shifting strategies can be implemented as precomputed indexing operations on the list holding the HEALPix features and are therefore efficient. In ablation studies we found that the spiral shifting outperforms the grid shifting slightly (see section~\ref{indoor}).

\subsubsection{Relative position bias}
In the SWIN-transformer, an important component that adds spatial information is the \emph{relative position bias} $B$ which is added to the query-key product in the attention layers: $\text{Att}(Q,K,V)=\text{SoftMax}(Q K^{\top}/\sqrt{d}+B)V\,.$
This bias is a learned value which depends only on the difference vector between the pixels in a pixel pair, i.e.\ all pixel pairs with the same relative position receive the same bias contribution.

In the HEALPix grid, the pixels inside each base quadrilateral are arranged in an approximately rectangular grid which we use to compute the relative positions of pixel pairs for obtaining the relative position bias mapping. Consequently, in HEAL-SWIN, $B\in\mathbb{R}^{\nwindow\times\nwindow}$ with values taken from a learnable matrix $\hat{B}\in\mathbb{R}^{(2\sqrt{\nwindow}-1)\times(2\sqrt{\nwindow}-1)}$ according to
\begin{align}
  B_{ij}=\hat{B}_{x(i)-x(j)+\sqrt{\nwindow},\ y(i)-y(j)+\sqrt{\nwindow}}\,,
\end{align}
where $(x(i),y(i))$ are the Cartesian coordinates of the pixel $i$ in the window, e.g.\ in Figure~\ref{fig:grid_shifting_scheme}, pixel 11 would have coordinates $x(11)=1$, $y(11)=0$. We share the same relative position bias table across all windows, so in particular also across base quadrilaterals. 
We also experimented with an absolute position embedding after the patch embedding layer but observed no benefit for performance.

\section{Experiments}
\label{sec:experiments}
To verify the performance of our model, we trained the HEAL-SWIN and the SWIN transformer on challenging realistic datasets of fisheye camera images of both street- and indoor scenes. We perform semantic segmentation and monocular depth estimation and furthermore test our model on the standard classification problem of spherical MNIST digits.

We show that HEAL-SWIN reaches better predictions than the non-spherical version of the model in all direct comparisons and is competitive on standard benchmarks for spherical models.

Due to space limitations, the details of the experiments on spherical MNIST are relegated to Appendix~\ref{app:additional_experiments}.

\subsection{Semantic segmentation of fisheye street scenes}
\label{subsec:sem_seg}
\begin{table}[t]
\centering
\caption{Mean intersection over union on the sphere for semantic segmentation of fisheye street scenes with HEAL-SWIN and SWIN, averaged over three runs.}
\label{tab:sem_seg_results}
\begin{tabular}{lll}
  \toprule
  Model & Dataset & mIoU \\
  \midrule
HEAL-SWIN & Large SynWoodScape & \textbf{0.947} \\ 
SWIN & Large SynWoodScape & 0.918 \\ 
\midrule
HEAL-SWIN & Large+AD SynWoodScape & \textbf{0.841} \\ 
SWIN & Large+AD SynWoodScape & 0.809 \\ 
\midrule
HEAL-SWIN & WoodScape & \textbf{0.628} \\ 
SWIN & WoodScape & 0.617 \\ 
\bottomrule
\end{tabular}
\end{table}

We compare the performance of HEAL-SWIN on HEALPix-projected fisheye images to that of the SWIN transformer on the original, i.e. flat and distorted images. The architecture and training hyperparameters were fixed by ablation studies for semantic segmentation on the WoodScape dataset unless stated otherwise.

As a baseline, we use the SWIN transformer in a $12$-layer configuration similar to the ``tiny'' configuration SWIN-T from the original paper~\cite{liu2021Swin} with a patch size of $2\times2$ and a window size of $8\times 8$, adapted to the size of our input images. Since both tasks require predictions of the same spatial dimensions as the input, we mirror the SWIN encoder in a SWIN decoder and add skip connections in a SWIN-UNet architecture, resulting in a model of around $41\mathrm{M}$ parameters. We found the improved layer-norm placement and cosine attention introduced in~\cite{liu2021swinv2} to be very effective and use them in all our models.

For the HEAL-SWIN models, we use the same configurations as for the SWIN model with a patch size of $\npatch=4$, mirroring the $2\times 2$ on the flat side and a window size of $\nwindow=64$, mirroring the $8\times 8$ on the flat side. For shifting, we use the spiral shifting introduced in Section~\ref{sec:shifting} with a shift size of $4$, corresponding roughly to half windows. Again, we mirror the encoder and add skip connections to obtain a UNet-like architecture. A table with the spatial feature dimensions throughout the network can be found in Appendix~\ref{app:heal-swin-model-details}. The shared model configuration gives our HEAL-SWIN model the same total parameter count as the SWIN model.

\subsubsection{Real-world images}
\label{sec:seg-real-world-images}
We evaluate our models using the real-world WoodScape dataset\footnote{For WoodScape, we noticed inconsistencies in the semantic labels, see Appendix~\ref{app:datasets-details} for further details.}~\cite{woodscape2019} consisting of 8234 fisheye images of street scenes recorded in various locations in the US, Europe and China. The images are presented as flat pixel grids together with calibration data which we use to project the input data and ground truth segmentation masks onto the sphere.

Although using only eight out of the twelve base pixel of the HEALPix grid allows for an efficient representation of the fisheye images, some image pixels are projected to regions outside of the coverage of our subset of the HEALPix grid; see the hatched regions of Figure~\ref{fig:half-sphere-image} in Appendix~\ref{app:datasets-details}. However, the affected pixels lie at the corners of the image, making the tradeoff well worth it for the autonomous driving tasks considered here. We restrict evaluation to the eight base pixels.

\begin{figure}
  \centering
  
  \includegraphics[width=0.40\textwidth]{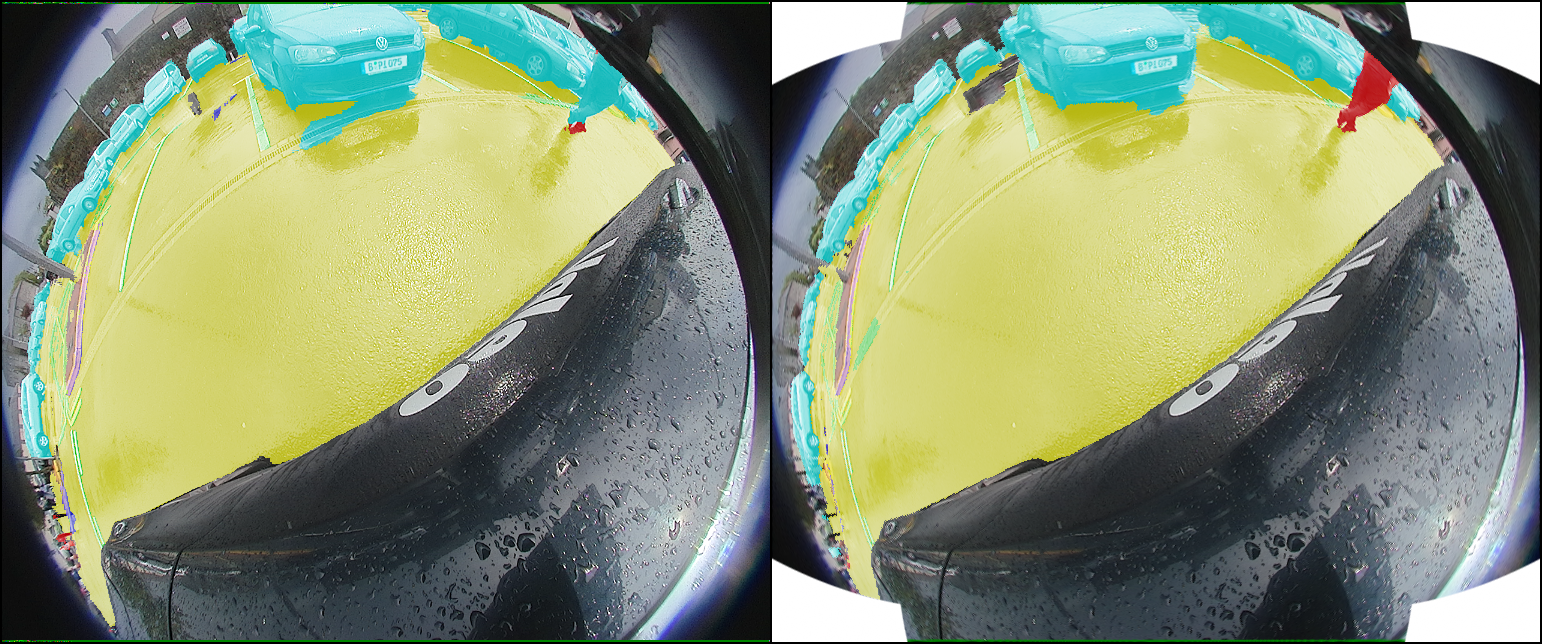}
  
  \caption{Example of segmentation using SWIN (left) and HEAL-SWIN (right) on the Woodscape dataset of real automotive images. Overlays correspond to predicted segmentation masks. The pedestrian (overlayed in red) is only recognized by HEAL-SWIN.}
  \label{fig:sample_predictions}
\end{figure}

For WoodScape, we exclude the \emph{void} class from the mean but keep it in the loss, different from the 2021 CVPR competition~\cite{ramachandran2021} as explained in Appendix~\ref{app:datasets-details}. As shown in Table~\ref{tab:sem_seg_results}, the HEALPix version of the SWIN transformer outperforms the baseline with otherwise identical training scheme and model architecture supporting the claimed benefit of spherical representations. See Figure \ref{fig:sample_predictions} for an example of the qualitative improvement of HEAL-SWIN over SWIN for the task of pedestrian segmentation. 

We investigated whether HEAL-SWIN outperforms the baseline in certain regions of the image and concluded that HEAL-SWIN outperforms SWIN everywhere and not just in a particular (e.g.\ very distorted) region of the image. This is likely the case since HEAL-SWIN can benefit from distorted and undistorted training regions equally, leading to a better overall performance.

\subsubsection{Synthetic images}
\label{sec:seg-synthetic-images}

\begin{table}
    \centering
    \caption{Inference times for semantic segmentation models on a single A40 GPU. Measurements are taken from first to last model operation in a forward pass with an input tensor of batch size one available on the GPU. Mean and standard deviation over 200 iterations with 10 iteration warm-up.}
    \label{tab:inference_times}
    \begin{tabular}{rlll}
      \toprule
      
      & Resolution & Pixels & time / pixel \\
      \midrule

      HEAL-SWIN & $8\times 256.0^2 $& $5.2\cdot 10^5$ & 297 $\pm$ 26ns \\
SWIN      & $640\times 768   $& $4.9\cdot 10^5$ & 296 $\pm$ 39ns \\
      \bottomrule
    \end{tabular}
\end{table}

In order to remove the effects of the labeling inaccuracies, we perform the analogous experiments with the same HEAL-SWIN model and SWIN baseline on the SynWoodScape~\cite{sekkat2022} dataset of 2000 fisheye images from synthetic street scenes generated using the driving simulator CARLA~\cite{carla17}. We use two different subsets out of the 25 classes provided in the dataset. All excluded classes are mapped to void. In the first subset, which we call \emph{Large SynWoodScape}, we train on 8 classes which cover large areas of the image, like \emph{building}, \emph{ego-vehicle}, \emph{road} etc.\ obtaining a dataset which lacks a lot of fine details and hence minimizes projection effects between the flat projection and HEALPix. For the second subset, \emph{Large+AD~SynWoodScape}, we include further classes relevant to autonomous driving, like \emph{pedestrian}, \emph{traffic light} and \emph{traffic sign} to create a more realistic dataset with 12 classes and featuring finer details; see Figure~\ref{fig:half-sphere-image}
in Appendix~\ref{app:datasets-details} for a sample. 
More details about the datasets 
are in Appendix~\ref{app:datasets-details}.

The performance results on the different datasets are summarized in Table~\ref{tab:sem_seg_results} and show that HEAL-SWIN outperforms the baseline in all cases. Figure~\ref{fig:main_results} (right) shows the results on \emph{Large+AD SynWoodScape}.

\paragraph{Inference time} 
Due to the efficient handling of the spherical data in HEAL-SWIN, the inference time is the same as the baseline model for the target resolution, cf. Table~\ref{tab:inference_times}. See Appendix Table \ref{tab:computational} for an ablation over resolutions. At lower resolutions, the memory layout in the HEALPix grid enables faster inference times for HEAL-SWIN compared to SWIN. For a comparison of the computational benefits of the SWIN architecture compared to CNNs we refer to~\cite{liu2021Swin}.

\begin{figure}
    \centering
    \includegraphics[width=0.7\linewidth]{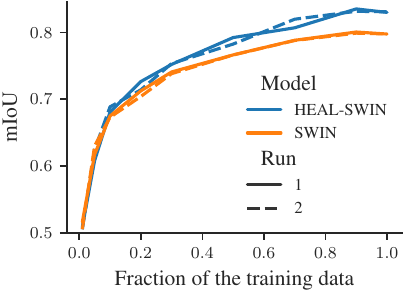}
    \caption{Semantic Segmentation for varying training set sizes. Performance is measured as the mean intersection over union (higher is better) computed on the HEALPix grid (spherical mIoU).}
    \label{fig:dataset_ablation_HEALPix}
    
\end{figure}
\paragraph{Dataset size ablation} We also study the effects of the size of the dataset by training on different subsets of the training data. We train on the \textit{Large+AD SynWoodScape} class subset. Within a run, we use exactly the same subset for both models, HEAL-SWIN and SWIN, while strictly increasing the subset when moving to a larger training set. All models are trained entirely from scratch until convergence and evaluated using the entire validation set. We find that the HEAL-SWIN model can make better use of larger training sets than the SWIN model, as the difference in performance becomes larger the more training data is used as shown in Figure~\ref{fig:dataset_ablation_HEALPix}. 

\subsection{Semantic segmentation of indoor fisheye images}
\label{indoor}
\begin{table}
  \centering
  \caption{Performance of spherical models on the Stanford 2D-3D segmentation task of indoor fisheye images evaluated using the official three-fold cross-validation.}
  \label{tab:stanford_dataset}
  \begin{tabular}{ccc}
    \toprule
    Model & mIoU & mAcc \\
    \midrule
    
    Gauge CNN~\cite{cohen2019gauge} & 39.4 & 55.9 \\
    UGSCNN~\cite{Jiang20219SphericalCNNUnstructuredGrids} & 38.8 & 54.7 \\
    HexRUNet~\cite{zhang2019b} & 43.3 & 58.6 \\
    SphCNN~\cite{esteves2018, esteves2020c} & 40.2 & 52.8 \\
    Spin-SphCNN~\cite{esteves2020c} & 41.9 & 55.6 \\
    HEAL-SWIN (Ours) & {\bf 44.3} & {\bf 61.9} \\
    \bottomrule
  \end{tabular}
\end{table}
To enable a comparison between our model and other models operating on spherical representations, we trained a version of HEAL-SWIN with about 1.5M parameters on the semantic masks of the Stanford 2D-3D-S dataset~\cite{Armeni2017Stanford} of 1413 RGB-D fisheye images of indoor scenes. We project the data to a HEALPix grid of resolution $\nside=64$, corresponding to $49\mathrm{k}$ pixels. For this task, we use all twelve base pixels of the grid. For details on the data, model architecture and training scheme, see Appendix~\ref{app:datasets-details} and \ref{app:heal-swin-model-details}.

In Table~\ref{tab:stanford_dataset}, we compare the performance of HEAL-SWIN to the performance of similarly-sized spherical models on the same dataset trained in a similar fashion (e.g.\ without data augmentation). HEAL-SWIN outperforms comparable models in this class. We use the same data preprocessing and weighted loss as~\cite{zhang2019b}. 

\paragraph{Ablations} To investigate how the model performance depends on the HEALPix-specific hyper parameters, we perform ablations over patch size, window size, shift size and shift strategy, as summarized in Appendix Table~\ref{tab:ablations}. The best model was obtained for window size $n_\mathrm{win}=16$, patch size $n_\mathrm{patch}=4$, and a shift size of $n_\mathrm{shift}=2$ with spiral shifting.

\subsection{Depth estimation}
\label{sec:depth-estimation}
\begin{figure}
    \centering
    \begin{minipage}{0.45\linewidth}
        \centering
        \includegraphics[height=0.8\linewidth]{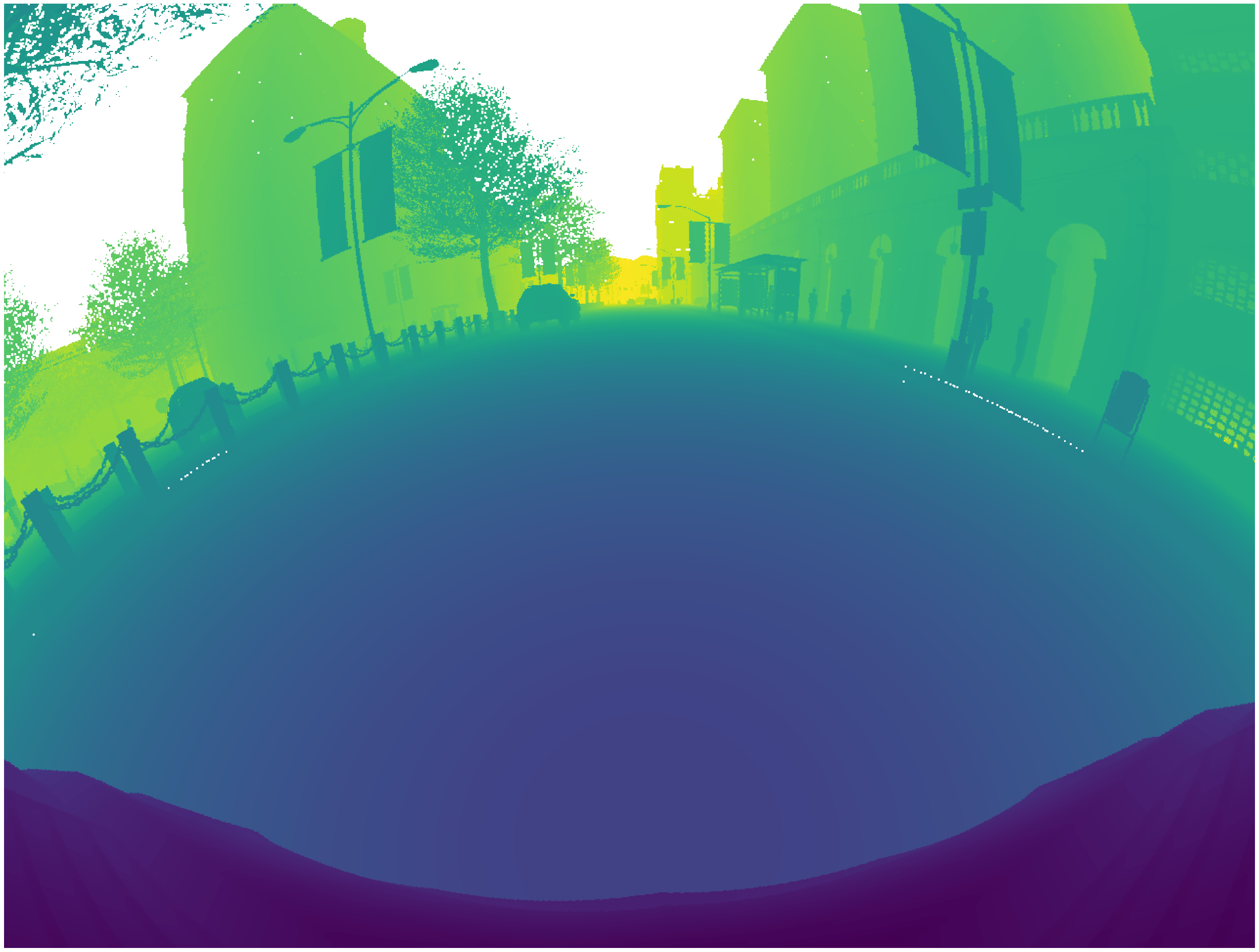}
    \end{minipage}
    \hspace*{\fill}
    \begin{minipage}{0.45\linewidth}
        \centering
        \begin{tikzpicture}
             \node at (0,0) {\includegraphics[trim={90 50 60 50}, clip, height=0.8\linewidth]{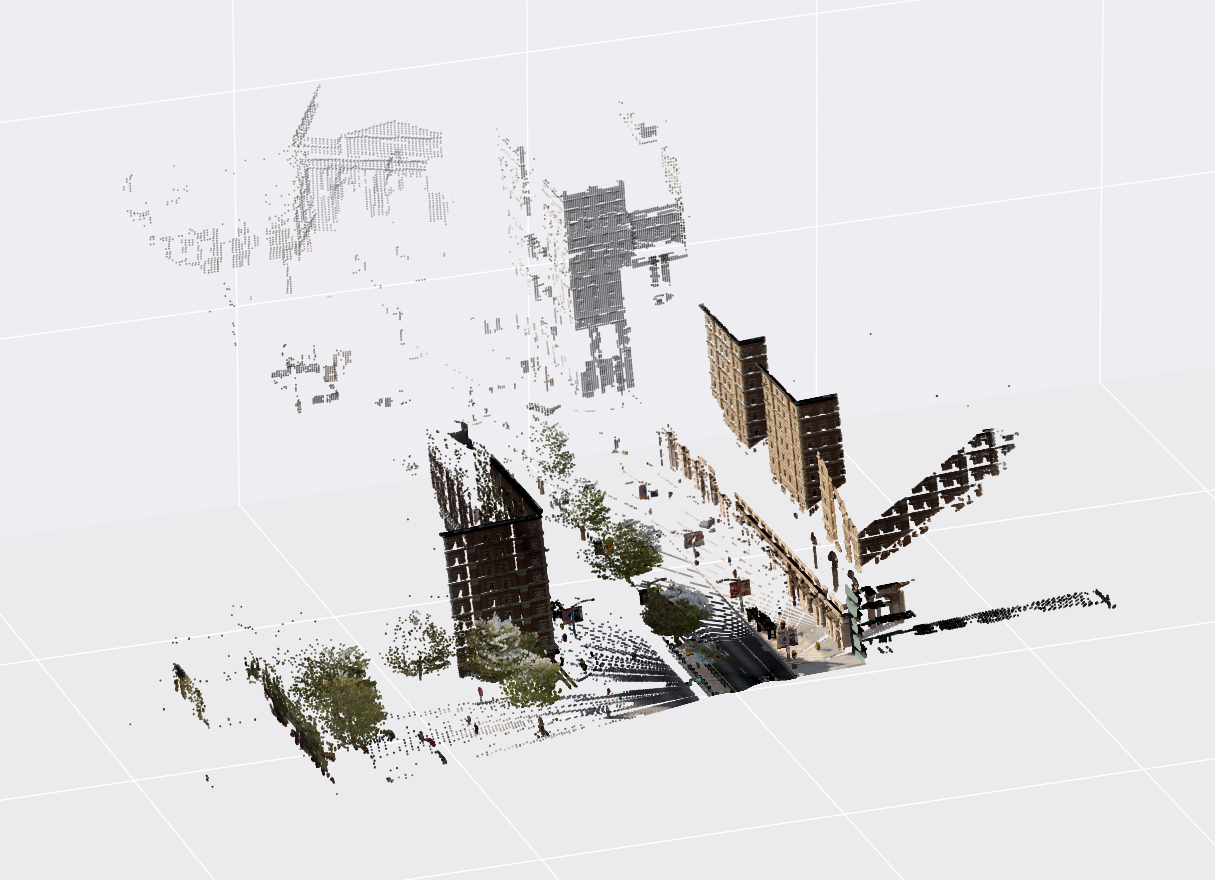}};
             \filldraw[color=red, fill=red, very thick](0.55,-1.0) circle (0.02); 
             \draw[color=red, ->, very thick] (0.55,-1.0) -- (0.38, -0.8);
        \end{tikzpicture}
    \end{minipage}
    \hspace*{\fill}
    \caption{
    Depth-map ground truth for the image in Appendix Figure~\ref{fig:half-sphere-image} (left) and corresponding point cloud (right). The red arrow indicates the orientation of the camera.
    }
    \label{fig:pointcloud}
\end{figure}

Estimating distances to obstacles and other road users is an important task for 3D scene understanding and route planning in autonomous driving. In monocular depth estimation, pixel-wise distance maps are predicted from camera images. For the SynWoodScape dataset, pixel-perfect depth maps are available on which we train our models. We use the same architectures and training procedures as in Section~\ref{subsec:sem_seg} and set the number of output channels to one.

The model is trained using an $L_2$ loss and the depth data is standardized to have zero mean and unit variance; in addition, the sky is masked out during training and evaluation. In order to preserve the common high-contrast edges all resampling is done using nearest neighbor interpolation.

In order to capture the quality of 3D scene predictions of the different models, we evaluate the depth estimations in terms of point clouds. More specifically, we generate a point cloud from the ground truth depth values by computing azimuthal and polar angles for each pixel from the calibration information of the camera and scaling the corresponding vectors on the unit sphere with the depth values, cf.\ Figure~\ref{fig:pointcloud} (right).  Similarly, the SWIN predictions are transformed into a point cloud, as are the HEAL-SWIN predictions, for which we use the pixel positions in HEALPix for the spherical angles. Comparing the predicted point clouds to the full-resolution ground truth point cloud leads to an evaluation scheme which is sensitive to the 3D information in the predicted depth values. These in turn are essential for downstream tasks. 

To compare the predicted point cloud $P_{\text{pred}}$ to the ground truth point cloud $P_{\text{gt}}$, we use the Chamfer distance~\cite{BarrowTenenbaum1977}.

From the results depicted in Figure~\ref{fig:main_results} (left),\footnote{For completeness, we want to mention that one HEAL-SWIN run performed very differently from the others with a Chamfer distance of $6.784$. According to Chauvenet's criterion this run should be classified as an outlier and is therefore not included in the figure.} it is evident that the point clouds predicted by HEAL-SWIN match the ground truth point cloud better than the point clouds predicted by the SWIN transformer, indicating that the HEAL-SWIN model has indeed learned a better 3D representation of the spherical images.

\section{Conclusion}

We constructed the efficient spherical vision tranformer HEAL-SWIN, combining the HEALPix spherical grid with the SWIN transformer. We showed superior performance of our model on the sphere, in comparison to a baseline SWIN model, for depth estimation and semantic segmentation on automotive and indoor fisheye images.

Although showing high performance already in its present form, HEAL-SWIN still has ample room for improvement. Firstly, in the presented setup, the grid is cut along base pixels to cover half of the sphere, leaving parts of the image uncovered while parts of the grid are unused. This could be improved by descending with the boundary into the nested structure of the grid and adapting the shifting strategy accordingly. Secondly, the relative position bias currently does not take into account the different base pixels around the poles and around the equator. This could be solved by a suitable correction deduced from the grid structure. Thirdly, the UNet-like architecture we base our setup on, is not state-of-the-art in tasks like semantic segmentation. Adapting a modern vision transformer decoder head to HEALPix could boost performance even further. 

Finally, our setup is not yet equivariant with respect to rotations of the sphere. For equivariant tasks like semantic segmentation or depth estimation, a considerable performance boost can be expected from making the model equivariant~\cite{gerken2022}. In this context it would also be very interesting to investigate equivariance with respect to local transformations. This has been thoroughly analyzed for CNNs in~\cite{cohen2019gauge,cheng2019covariance, gerken2021geometric} and a gauge equivariant transformer has been proposed in~\cite{He2021GaugeET}.

\section*{Acknowledgments and Disclosure of Funding}
 \noindent We are very grateful to Jimmy Aronsson and Joakim Johnander for discussions, feedback on the manuscript and for collaborations in the initial stages of this project. 

 The work of O.C., J.G. and D.P. is supported by the Wallenberg AI, Autonomous Systems and Software Program
(WASP) funded by the Knut and Alice Wallenberg Foundation. J.G. was supported by the Berlin Institute for the Foundations of Learning and Data (BIFOLD). H.S. work has been funded by the Deutsche Forschungsgemeinschaft (DFG, German Research Foundation) under Germany’s Excellence Strategy – EXC 2002/1 “Science of Intelligence” – project number 390523135. The computations were enabled by resources provided by the National Academic Infrastructure for Supercomputing in Sweden (NAISS) and the Swedish National Infrastructure for Computing (SNIC) at C3SE partially funded by the Swedish Research Council through grant agreements 2022-06725 and 2018-05973.

\small
\bibliography{cites}
\bibliographystyle{ieeenat_fullname}

\clearpage
\appendix
\maketitlesupplementary

\section{Datasets}
\label{app:datasets-details}
\paragraph{WoodScape and SynWoodScape} For our experiments we use the WoodScape~\cite{woodscape2019} and the SynWoodScape~\cite{sekkat2022} dataset. Note that we used the 2k samples which were published at the time of writing\footnote{\url{https://drive.google.com/drive/folders/1N5rrySiw1uh9kLeBuOblMbXJ09YsqO7I}}, instead of the full 80k samples. In all experiments, we split the available samples randomly (but consistently across models and runs) into 80\

\begin{table*}[b!]
  \centering
  \caption{Classes from the Large SynWoodScape and the Large+AD SynWoodScape datasets in terms of the classes provided by SynWoodScape.}
  \vspace{\baselineskip}
  \begin{tabular}{ccc}
    \toprule
    \multirow{2}{*}{SynWoodScape} & \multicolumn{2}{c}{Our Classes}\\
    \cmidrule(lr){2-3}
    & Large SynWoodScape & Large+AD SynWoodScape\\
    \midrule
    unlabeled & void & void \\
    building & building & building \\
    fence & void & void \\
    other & void & void \\
    pedestrian & void & pedestrian \\
    pole & void & void \\
    road line & road line & road line\\
    road & road & road \\
    sidewalk & sidewalk & sidewalk \\
    vegetation & void & void \\
    four-wheeler vehicle & four-wheeler vehicle & four-wheeler vehicle \\
    wall & void & void \\
    traffic sign & void & traffic sign \\
    sky & sky & sky \\
    ground & void & void \\
    bridge & void & void \\
    rail track & void & void \\
    guard rail & void & void \\
    traffic light & void & traffic light \\
    water & void & void \\
    terrain & void & void \\
    two-wheeler vehicle & void & two-wheeler vehicle \\
    static & void & void \\
    dynamic & void & void \\
    ego-vehicle & ego-vehicle & ego-vehicle \\
    \bottomrule
  \end{tabular}
  \label{tab:subset_classes}
\end{table*}

In the 2021 CVPR competition for segmentation of the WoodScape dataset~\cite{ramachandran2021}, pixels that are labeled with the dominant \emph{void} class in the ground truth were excluded from the mIoU used for ranking. Therefore, many teams excluded the void class from their training loss, resulting in random predictions for large parts of the image. This shortcoming was noted in~\cite{ramachandran2021}, but the evaluation score could not be changed after the competition had been published. Given these circumstances, we decided to include the \emph{void} class into our training loss but exclude it from the mean over classes in the mIoU to more accurately reflect the performance of our models on the more difficult classes. However, this also means that our results cannot be directly compared to the results of the competition.

The same problem does not arise for the SynWoodScape dataset and our two variants since their class lists include all major structures in the image, leading to a much reduced prevalence of the void class. Therefore, we include all classes in the mIoU for these datasets.

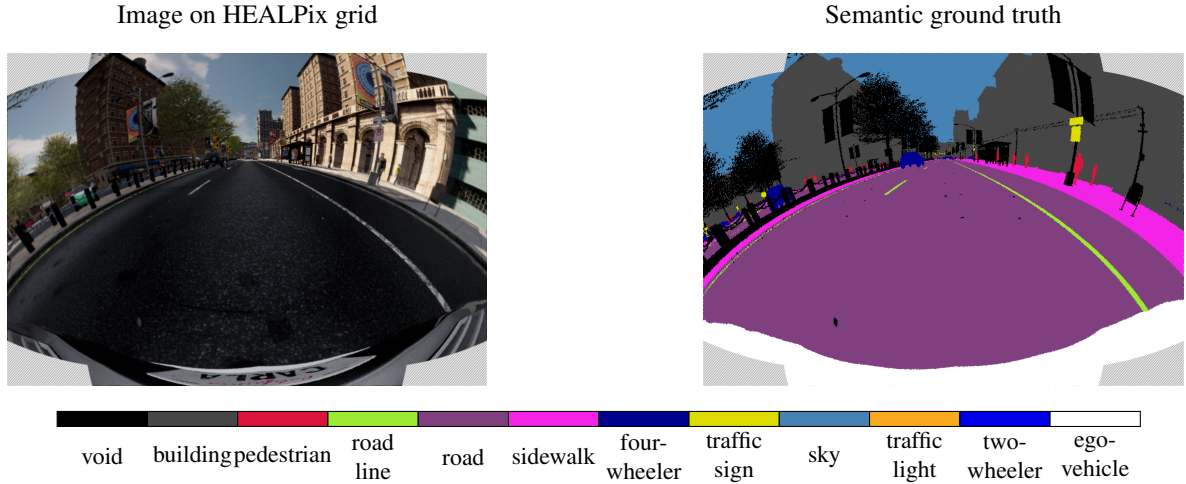
\begin{figure*}
  \centering
  \begin{minipage}{0.47\linewidth}
  \centering
  \begin{tikzpicture}

\definecolor{darkgray176}{RGB}{176,176,176}

\begin{axis}[
height=1.5\figwh,
width=1.5\figww,
hide x axis,
hide y axis,
tick align=outside,
tick pos=left,
title={Image on HEALPix grid},
x grid style={darkgray176},
xmin=-0.5, xmax=1279.5,
xtick style={color=black},
y dir=reverse,
y grid style={darkgray176},
ymin=-0.5, ymax=965.5,
ytick style={color=black}
]
\addplot graphics [includegraphics cmd=\pgfimage,xmin=-0.5, xmax=1279.5, ymin=965.5, ymax=-0.5] {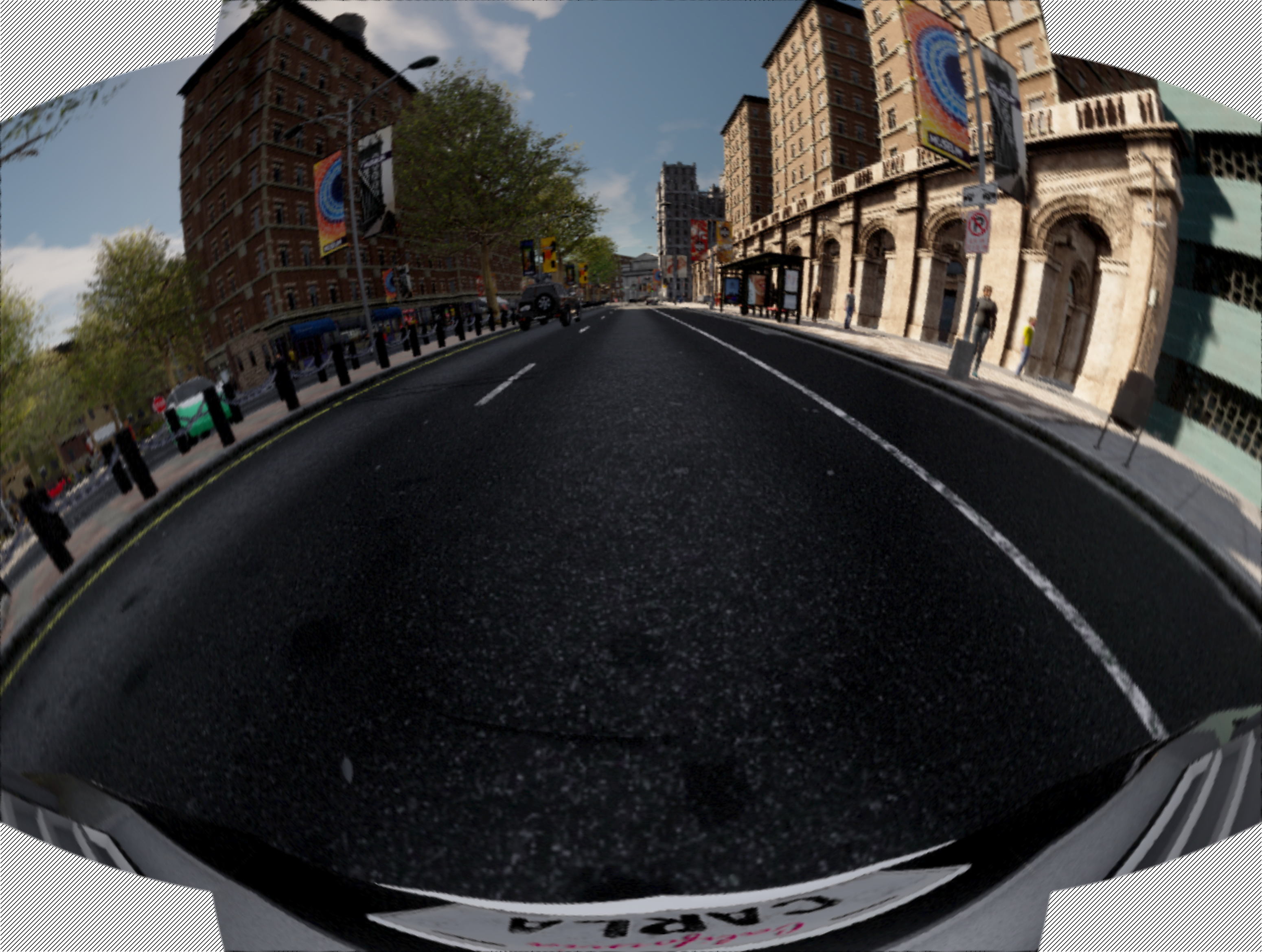};
\end{axis}

\end{tikzpicture}
  \end{minipage}
  \hspace*{\fill}
  \begin{minipage}{0.47\linewidth}
  \centering
  \begin{tikzpicture}

\definecolor{darkgray176}{RGB}{176,176,176}

\begin{axis}[
height=1.5\figwh,
width=1.5\figww,
hide x axis,
hide y axis,
tick align=outside,
tick pos=left,
title={Semantic ground truth},
x grid style={darkgray176},
xmin=-0.5, xmax=1279.5,
xtick style={color=black},
y dir=reverse,
y grid style={darkgray176},
ymin=-0.5, ymax=965.5,
ytick style={color=black}
]
\addplot graphics [includegraphics cmd=\pgfimage,xmin=-0.5, xmax=1279.5, ymin=965.5, ymax=-0.5] {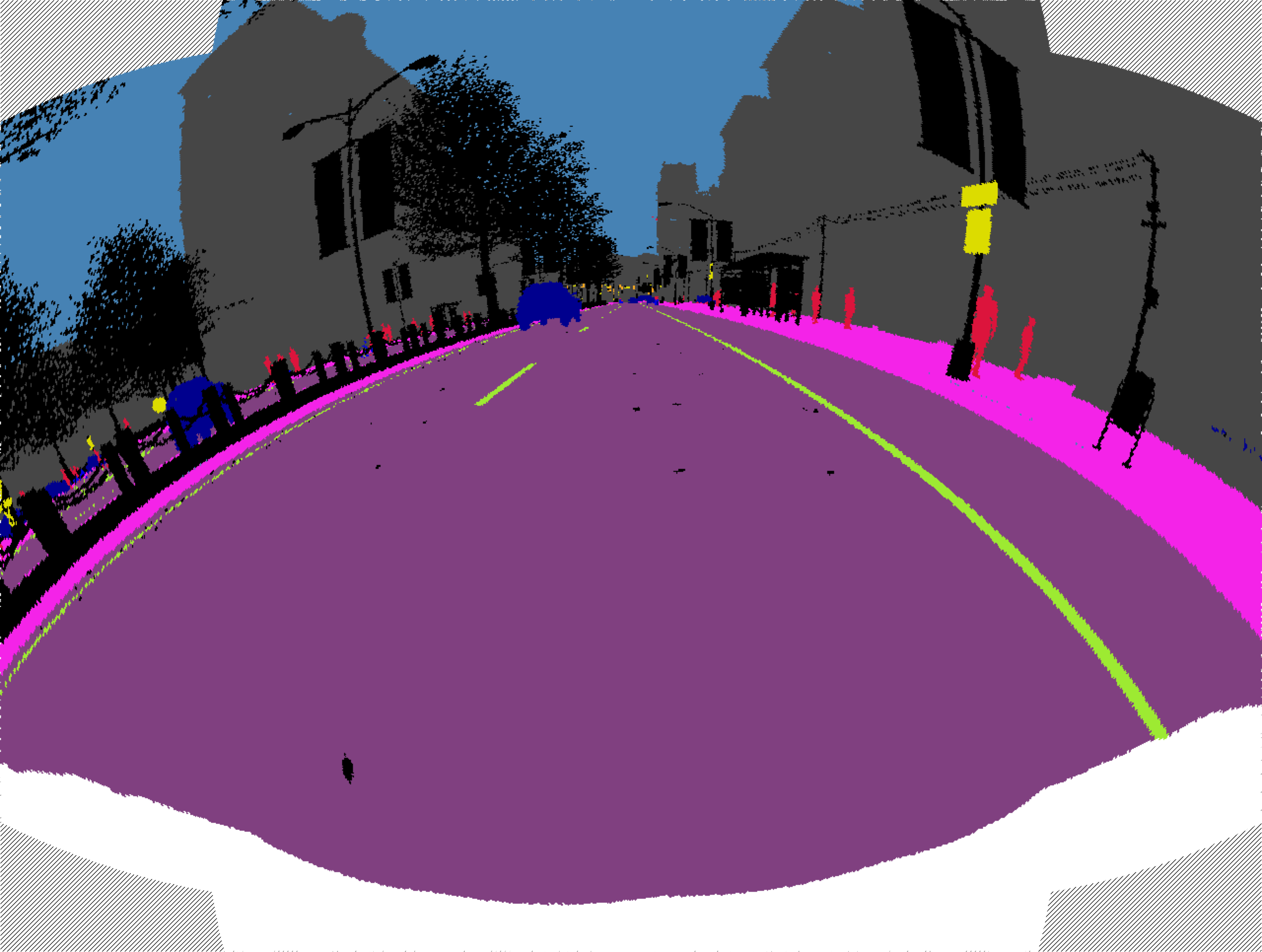};
\end{axis}

\end{tikzpicture}
  \end{minipage}\\

  \vspace*{0.3cm}
  \definecolor{darkslategray38}{RGB}{38,38,38}
\definecolor{gray}{RGB}{128,128,128}
\definecolor{lavender234234242}{RGB}{234,234,242}
\definecolor{steelblue76114176}{RGB}{76,114,176}

\definecolor{colvoi}{RGB}{0, 0, 0}
\definecolor{colbui}{RGB}{70, 70, 70}
\definecolor{colped}{RGB}{220, 20, 60}
\definecolor{colrol}{RGB}{157, 234, 50}
\definecolor{colroa}{RGB}{128, 64, 128}
\definecolor{colsid}{RGB}{244, 35, 232}
\definecolor{colfou}{RGB}{0, 0, 142}
\definecolor{coltrs}{RGB}{220, 220, 0}
\definecolor{colsky}{RGB}{70, 130, 180}
\definecolor{coltrl}{RGB}{250, 170, 30}
\definecolor{coltwo}{RGB}{0, 0, 230}
\definecolor{colego}{RGB}{255, 255, 255}

\newcommand{\recheight}{0.2}
\newcommand{\recspace}{1.2}
\newcommand{\recwidth}{\recspace}
\newcommand{\ybase}{5.5}
\newcommand{\ysep}{0.4}

\begin{tikzpicture}
  \foreach \ind/\col/\lab in {
         0/colvoi/void,
         1/colbui/building,
         2/colped/pedestrian,
         3/colrol/road\\line,
         4/colroa/road,
         5/colsid/sidewalk,
         6/colfou/four-\\wheeler,
         7/coltrs/traffic\\sign,
         8/colsky/sky,
         9/coltrl/traffic\\light,
        10/coltwo/two-\\wheeler,
        11/colego/ego-\\vehicle
    } {
    \draw[fill=\col,draw=black] (0+\recspace*\ind,\ybase) rectangle ++(\recwidth,\recheight);
    \node[font=\small,align=center] at (0.5*\recspace + \recspace*\ind,\ybase - \ysep) {\lab};
    }
\end{tikzpicture}
  \caption{Sample RGB image (left) and semantic segmentation ground truth (right) from the Large+AD SynWoodScape dataset, projected onto the plane for visualization. Regions not covered by the $\sfrac{8}{12}$ base pixels of the HEALPix grid are hatched.}
  \label{fig:half-sphere-image}
  \vspace{-\baselineskip}
\end{figure*}

\begin{figure*}[p]
  \centering
  \begin{tikzpicture}

\definecolor{color0}{rgb}{0.917647058823529,0.917647058823529,0.949019607843137}
\definecolor{color1}{rgb}{0.298039215686275,0.447058823529412,0.690196078431373}

\begin{axis}[
width=\textwidth,
height=0.18\textheight,
title={WoodScape},
axis background/.style={fill=color0},
axis line style={white},
tick align=outside,
tick pos=left,
x grid style={white},
xmajorgrids,
xmin=-0.6, xmax=9.6,
xtick style={color=white!15!black},
xtick={0,1,2,3,4,5,6,7,8,9},
xticklabel style={rotate=45.0,anchor=east},
xticklabels={
   void,
   road,
   lanemarks,
   curb,
   person,
   rider,
   vehicles,
   bicycle,
   motorcycle,
   traffic\_sign
},
y grid style={white},
ylabel={Percent of total pixels},
ymajorgrids,
ymin=0, ymax=70.9696483612061,
ytick style={color=white!15!black}
]
\draw[draw=none,fill=color1,line width=0.12pt] (axis cs:-0.4,0) rectangle (axis cs:0.4,67.5901412963867);
\draw[draw=none,fill=color1,line width=0.12pt] (axis cs:0.6,0) rectangle (axis cs:1.4,25.8006153106689);
\draw[draw=none,fill=color1,line width=0.12pt] (axis cs:1.6,0) rectangle (axis cs:2.4,1.52900874614716);
\draw[draw=none,fill=color1,line width=0.12pt] (axis cs:2.6,0) rectangle (axis cs:3.4,0.562553346157074);
\draw[draw=none,fill=color1,line width=0.12pt] (axis cs:3.6,0) rectangle (axis cs:4.4,0.32955938577652);
\draw[draw=none,fill=color1,line width=0.12pt] (axis cs:4.6,0) rectangle (axis cs:5.4,0.118134789168835);
\draw[draw=none,fill=color1,line width=0.12pt] (axis cs:5.6,0) rectangle (axis cs:6.4,3.49804711341858);
\draw[draw=none,fill=color1,line width=0.12pt] (axis cs:6.6,0) rectangle (axis cs:7.4,0.416775107383728);
\draw[draw=none,fill=color1,line width=0.12pt] (axis cs:7.6,0) rectangle (axis cs:8.4,0.123026743531227);
\draw[draw=none,fill=color1,line width=0.12pt] (axis cs:8.6,0) rectangle (axis cs:9.4,0.0321342684328556);
\end{axis}

\end{tikzpicture}
  \begin{tikzpicture}

\definecolor{color0}{rgb}{0.917647058823529,0.917647058823529,0.949019607843137}
\definecolor{color1}{rgb}{0.298039215686275,0.447058823529412,0.690196078431373}

\begin{axis}[
width=\textwidth,
height=0.18\textheight,
title={SynWoodScape},
axis background/.style={fill=color0},
axis line style={white},
tick align=outside,
tick pos=left,
x grid style={white},
xmajorgrids,
xmin=-0.6, xmax=24.6,
xtick style={color=white!15!black},
xtick={0,1,2,3,4,5,6,7,8,9,10,11,12,13,14,15,16,17,18,19,20,21,22,23,24},
xticklabel style={rotate=38.0,anchor=east},
xticklabels={
   unlabeled,
   building,
   fence,
   other,
   pedestrian,
   pole,
   road line,
   road,
   sidewalk,
   vegetation,
   four-wheeler vehicle,
   wall,
   traffic sign,
   sky,
   ground,
   bridge,
   rail track,
   guard rail,
   traffic light,
   water,
   terrain,
   two-wheeler vehicle,
   static,
   dynamic,
   ego-vehicle
},
y grid style={white},
ylabel={Percent of total pixels},
ymajorgrids,
ymin=0, ymax=41.266473197937,
ytick style={color=white!15!black}
]
\draw[draw=none,fill=color1,line width=0.12pt] (axis cs:-0.4,0) rectangle (axis cs:0.4,0);
\draw[draw=none,fill=color1,line width=0.12pt] (axis cs:0.6,0) rectangle (axis cs:1.4,14.7425832748413);
\draw[draw=none,fill=color1,line width=0.12pt] (axis cs:1.6,0) rectangle (axis cs:2.4,0.165106356143951);
\draw[draw=none,fill=color1,line width=0.12pt] (axis cs:2.6,0) rectangle (axis cs:3.4,0.879262208938599);
\draw[draw=none,fill=color1,line width=0.12pt] (axis cs:3.6,0) rectangle (axis cs:4.4,0.519136965274811);
\draw[draw=none,fill=color1,line width=0.12pt] (axis cs:4.6,0) rectangle (axis cs:5.4,0.460873186588287);
\draw[draw=none,fill=color1,line width=0.12pt] (axis cs:5.6,0) rectangle (axis cs:6.4,1.60426306724548);
\draw[draw=none,fill=color1,line width=0.12pt] (axis cs:6.6,0) rectangle (axis cs:7.4,39.3014030456543);
\draw[draw=none,fill=color1,line width=0.12pt] (axis cs:7.6,0) rectangle (axis cs:8.4,3.20932722091675);
\draw[draw=none,fill=color1,line width=0.12pt] (axis cs:8.6,0) rectangle (axis cs:9.4,3.09217166900635);
\draw[draw=none,fill=color1,line width=0.12pt] (axis cs:9.6,0) rectangle (axis cs:10.4,1.63187444210052);
\draw[draw=none,fill=color1,line width=0.12pt] (axis cs:10.6,0) rectangle (axis cs:11.4,0.088720440864563);
\draw[draw=none,fill=color1,line width=0.12pt] (axis cs:11.6,0) rectangle (axis cs:12.4,0.0739461630582809);
\draw[draw=none,fill=color1,line width=0.12pt] (axis cs:12.6,0) rectangle (axis cs:13.4,4.86802339553833);
\draw[draw=none,fill=color1,line width=0.12pt] (axis cs:13.6,0) rectangle (axis cs:14.4,0.130118921399117);
\draw[draw=none,fill=color1,line width=0.12pt] (axis cs:14.6,0) rectangle (axis cs:15.4,0.000823668786324561);
\draw[draw=none,fill=color1,line width=0.12pt] (axis cs:15.6,0) rectangle (axis cs:16.4,0.106121316552162);
\draw[draw=none,fill=color1,line width=0.12pt] (axis cs:16.6,0) rectangle (axis cs:17.4,0);
\draw[draw=none,fill=color1,line width=0.12pt] (axis cs:17.6,0) rectangle (axis cs:18.4,0.0333744995296001);
\draw[draw=none,fill=color1,line width=0.12pt] (axis cs:18.6,0) rectangle (axis cs:19.4,0.00326006091199815);
\draw[draw=none,fill=color1,line width=0.12pt] (axis cs:19.6,0) rectangle (axis cs:20.4,0.0228742882609367);
\draw[draw=none,fill=color1,line width=0.12pt] (axis cs:20.6,0) rectangle (axis cs:21.4,0.119342721998692);
\draw[draw=none,fill=color1,line width=0.12pt] (axis cs:21.6,0) rectangle (axis cs:22.4,0.670351684093475);
\draw[draw=none,fill=color1,line width=0.12pt] (axis cs:22.6,0) rectangle (axis cs:23.4,0);
\draw[draw=none,fill=color1,line width=0.12pt] (axis cs:23.6,0) rectangle (axis cs:24.4,28.2770385742188);
\end{axis}

\end{tikzpicture}
  \begin{tikzpicture}

\definecolor{color0}{rgb}{0.917647058823529,0.917647058823529,0.949019607843137}
\definecolor{color1}{rgb}{0.298039215686275,0.447058823529412,0.690196078431373}

\begin{axis}[
width=\textwidth,
height=0.18\textheight,
title={Large SynWoodScape},
axis background/.style={fill=color0},
axis line style={white},
tick align=outside,
tick pos=left,
x grid style={white},
xmajorgrids,
xmin=-0.6, xmax=7.6,
xtick style={color=white!15!black},
xtick={0,1,2,3,4,5,6,7},
xticklabel style={rotate=38.0,anchor=east},
xticklabels={
   void,
   building,
   road line,
   road,
   sidewalk,
   four-wheeler vehicle,
   sky,
   ego-vehicle
},
y grid style={white},
ylabel={Percent of total pixels},
ymajorgrids,
ymin=0, ymax=41.266473197937,
ytick style={color=white!15!black}
]
\draw[draw=none,fill=color1,line width=0.12pt] (axis cs:-0.4,0) rectangle (axis cs:0.4,6.3654842376709);
\draw[draw=none,fill=color1,line width=0.12pt] (axis cs:0.6,0) rectangle (axis cs:1.4,14.7425832748413);
\draw[draw=none,fill=color1,line width=0.12pt] (axis cs:1.6,0) rectangle (axis cs:2.4,1.60426306724548);
\draw[draw=none,fill=color1,line width=0.12pt] (axis cs:2.6,0) rectangle (axis cs:3.4,39.3014030456543);
\draw[draw=none,fill=color1,line width=0.12pt] (axis cs:3.6,0) rectangle (axis cs:4.4,3.20932722091675);
\draw[draw=none,fill=color1,line width=0.12pt] (axis cs:4.6,0) rectangle (axis cs:5.4,1.63187444210052);
\draw[draw=none,fill=color1,line width=0.12pt] (axis cs:5.6,0) rectangle (axis cs:6.4,4.86802339553833);
\draw[draw=none,fill=color1,line width=0.12pt] (axis cs:6.6,0) rectangle (axis cs:7.4,28.2770385742188);
\end{axis}

\end{tikzpicture}
  \begin{tikzpicture}

\definecolor{color0}{rgb}{0.917647058823529,0.917647058823529,0.949019607843137}
\definecolor{color1}{rgb}{0.298039215686275,0.447058823529412,0.690196078431373}

\begin{axis}[
width=\textwidth,
height=0.18\textheight,
title={Large+AD SynWoodScape},
axis background/.style={fill=color0},
axis line style={white},
tick align=outside,
tick pos=left,
x grid style={white},
xmajorgrids,
xmin=-0.6, xmax=11.6,
xtick style={color=white!15!black},
xtick={0,1,2,3,4,5,6,7,8,9,10,11},
xticklabel style={rotate=38.0,anchor=east},
xticklabels={
   void,
   building,
   pedestrian,
   road line,
   road,
   sidewalk,
   four-wheeler vehicle,
   traffic sign,
   sky,
   traffic light,
   two-wheeler vehicle,
   ego-vehicle
},
y grid style={white},
ylabel={Percent of total pixels},
ymajorgrids,
ymin=0, ymax=41.266473197937,
ytick style={color=white!15!black}
]
\draw[draw=none,fill=color1,line width=0.12pt] (axis cs:-0.4,0) rectangle (axis cs:0.4,5.61968421936035);
\draw[draw=none,fill=color1,line width=0.12pt] (axis cs:0.6,0) rectangle (axis cs:1.4,14.7425832748413);
\draw[draw=none,fill=color1,line width=0.12pt] (axis cs:1.6,0) rectangle (axis cs:2.4,0.519136965274811);
\draw[draw=none,fill=color1,line width=0.12pt] (axis cs:2.6,0) rectangle (axis cs:3.4,1.60426306724548);
\draw[draw=none,fill=color1,line width=0.12pt] (axis cs:3.6,0) rectangle (axis cs:4.4,39.3014030456543);
\draw[draw=none,fill=color1,line width=0.12pt] (axis cs:4.6,0) rectangle (axis cs:5.4,3.20932722091675);
\draw[draw=none,fill=color1,line width=0.12pt] (axis cs:5.6,0) rectangle (axis cs:6.4,1.63187444210052);
\draw[draw=none,fill=color1,line width=0.12pt] (axis cs:6.6,0) rectangle (axis cs:7.4,0.0739461630582809);
\draw[draw=none,fill=color1,line width=0.12pt] (axis cs:7.6,0) rectangle (axis cs:8.4,4.86802339553833);
\draw[draw=none,fill=color1,line width=0.12pt] (axis cs:8.6,0) rectangle (axis cs:9.4,0.0333744995296001);
\draw[draw=none,fill=color1,line width=0.12pt] (axis cs:9.6,0) rectangle (axis cs:10.4,0.119342721998692);
\draw[draw=none,fill=color1,line width=0.12pt] (axis cs:10.6,0) rectangle (axis cs:11.4,28.2770385742188);
\end{axis}

\end{tikzpicture}
  \caption{Class distributions for the datasets used in semantic segmentation.}
  \label{fig:class_hists}
\end{figure*}
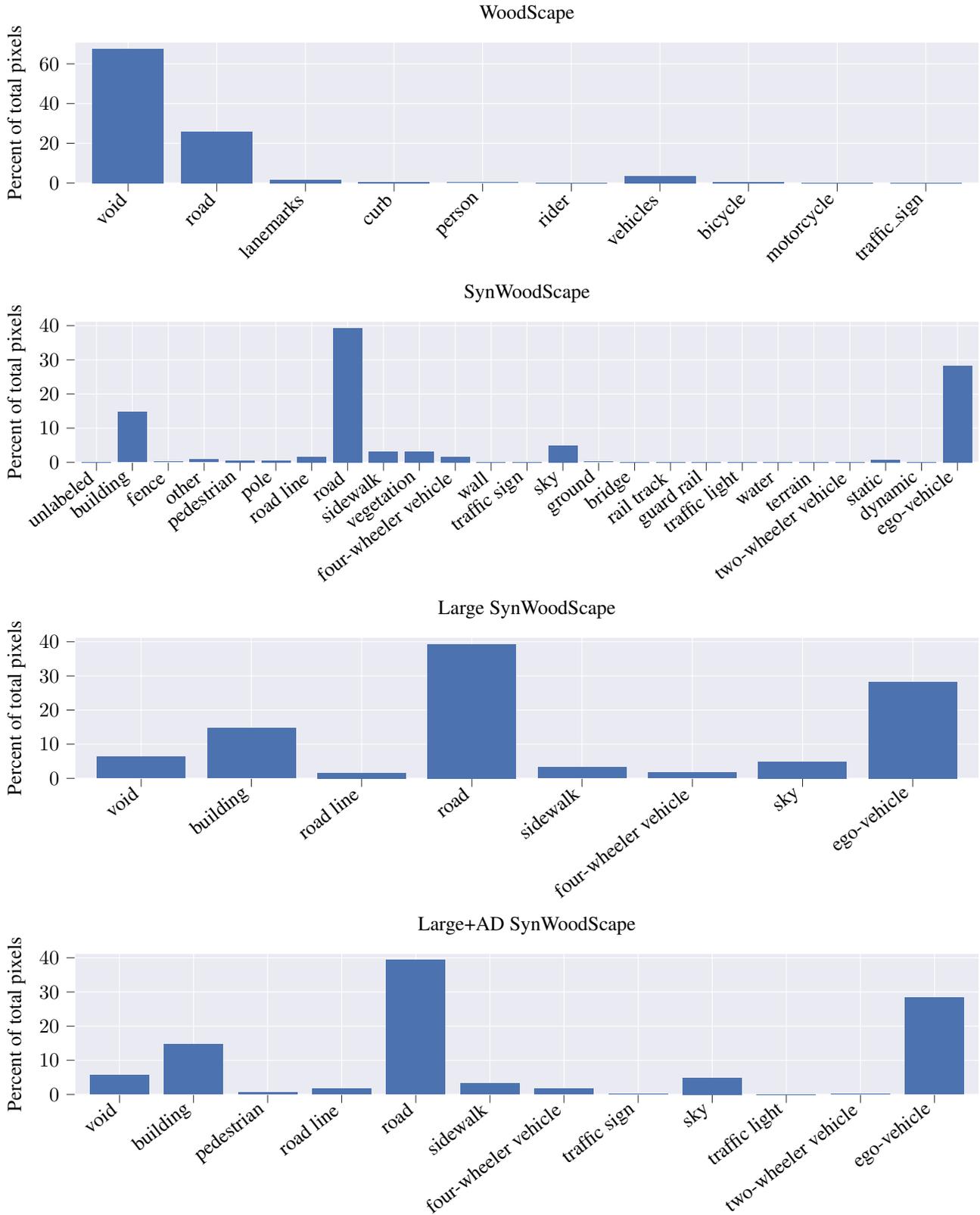

\begin{figure*}[p]
  \centering
  \begin{subfigure}{\textwidth}
    \centering
    \includegraphics[width=0.35\textwidth]{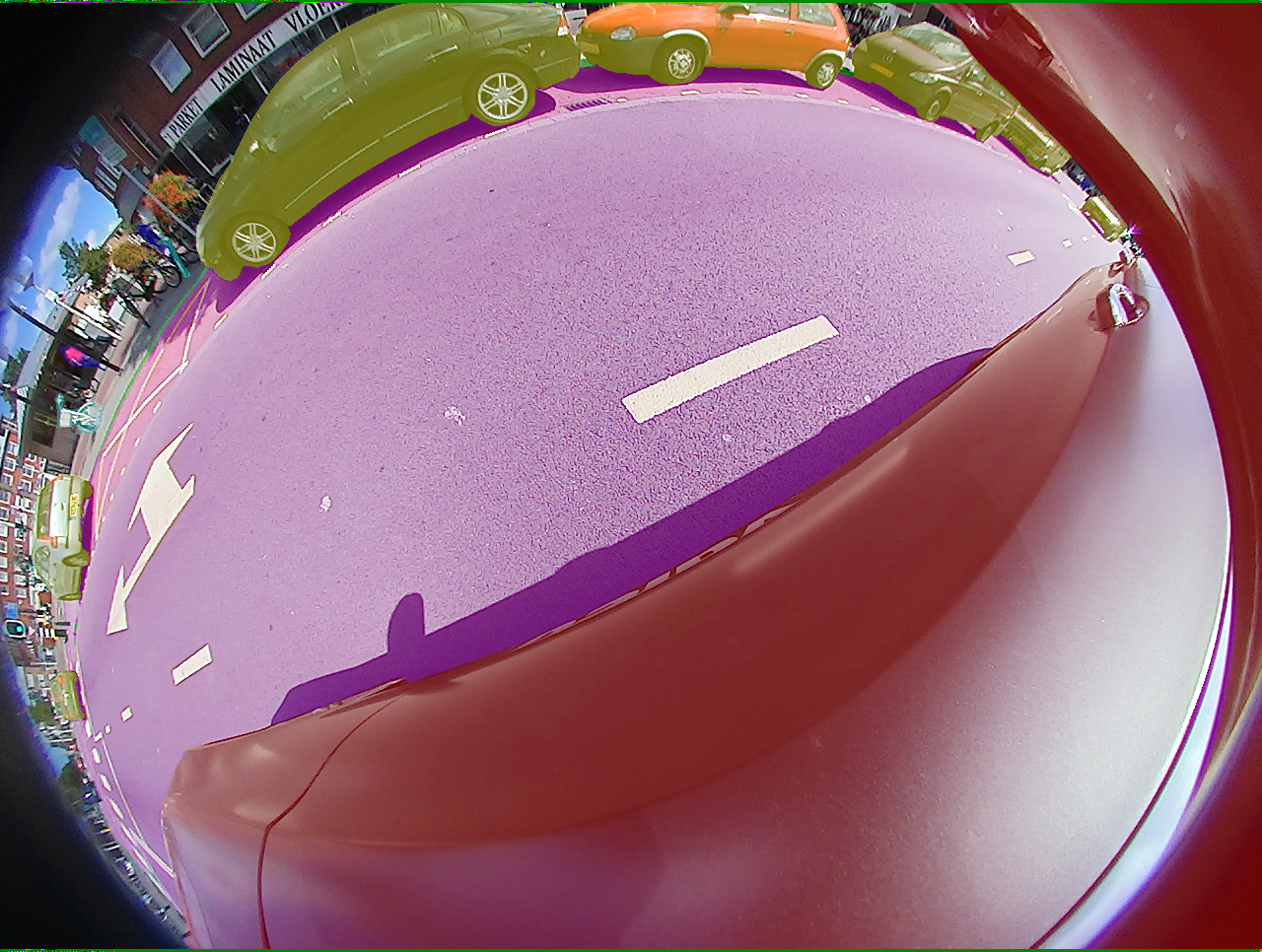}
    \caption{Large parts of the ego vehicle are labeled as \emph{lanemarks}.}
    \vspace{\baselineskip}
  \end{subfigure}
  \begin{subfigure}{\textwidth}
    \centering
    \includegraphics[width=0.35\textwidth]{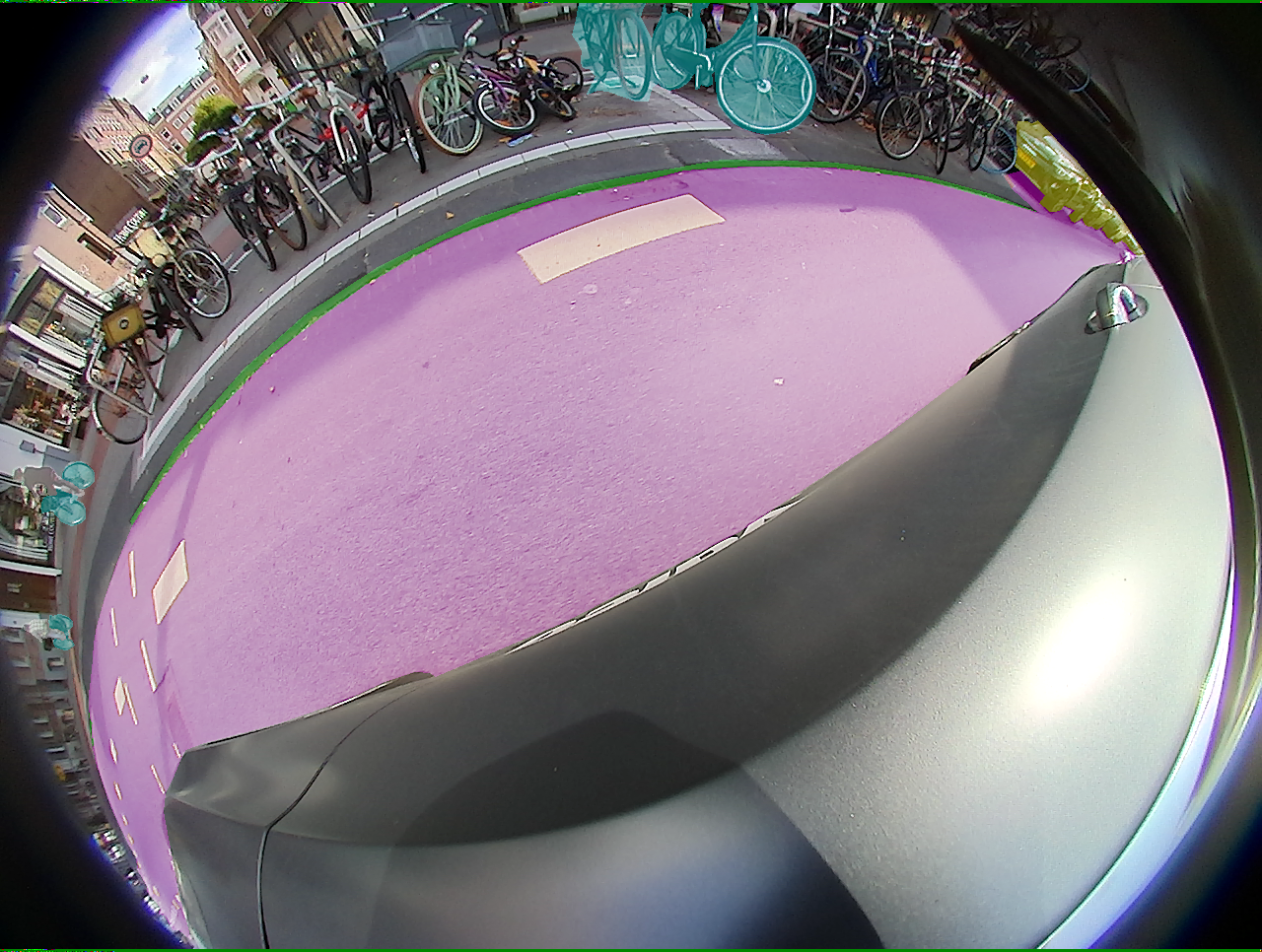}
    \hspace{1em}
    \includegraphics[width=0.35\textwidth]{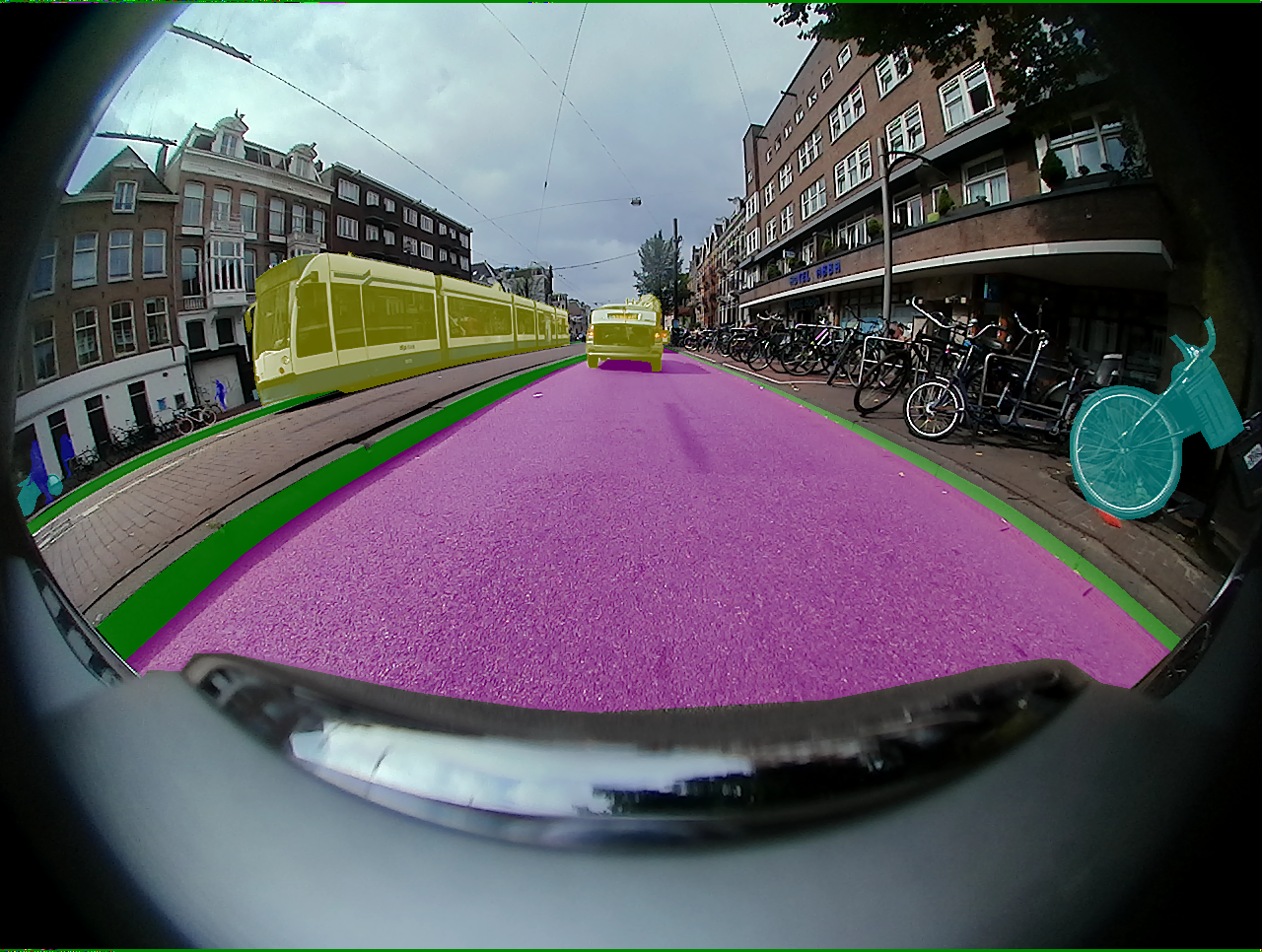}\\[0.5ex]
    \includegraphics[width=0.35\textwidth]{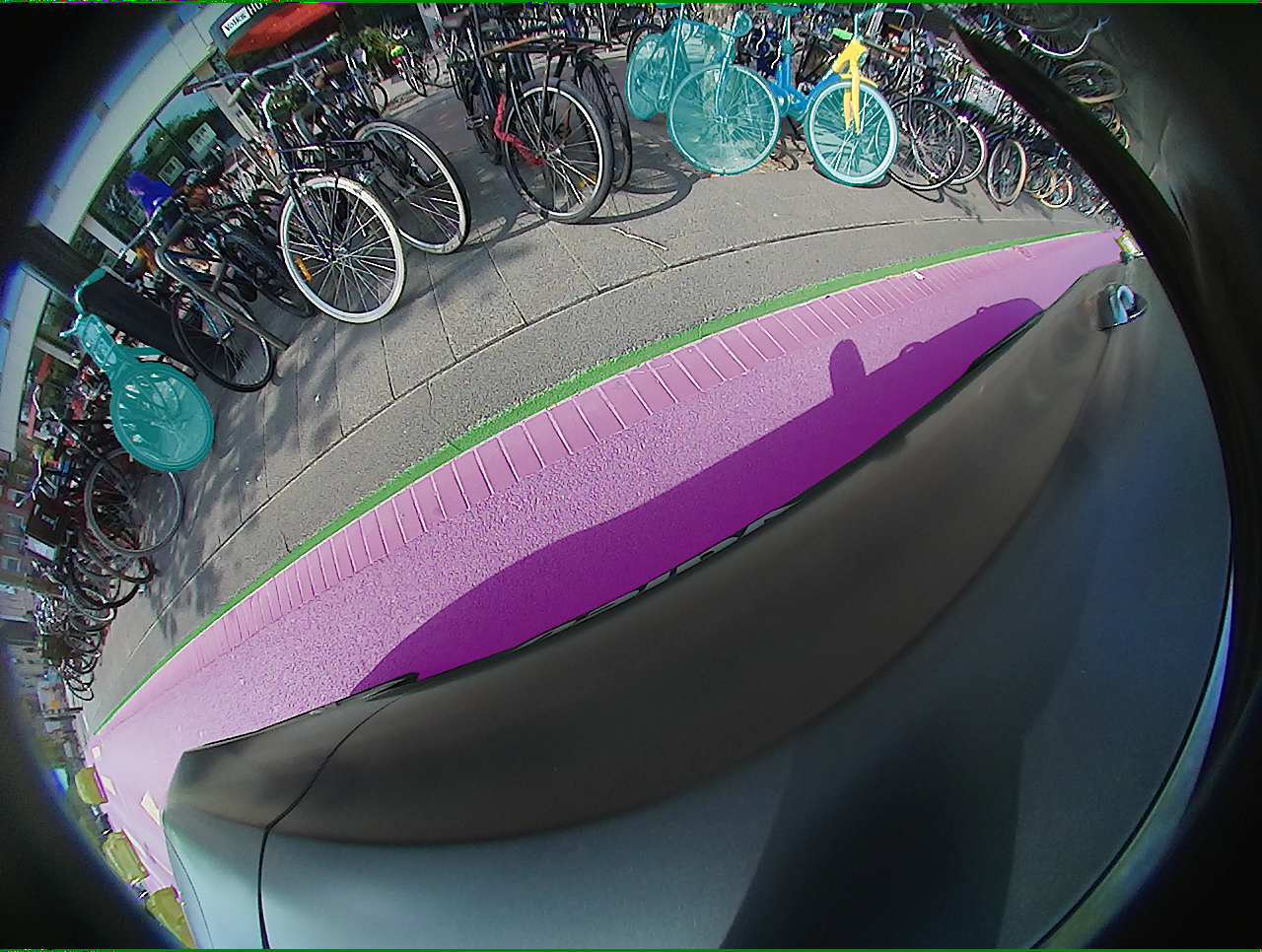}
    \hspace{1em}
    \includegraphics[width=0.35\textwidth]{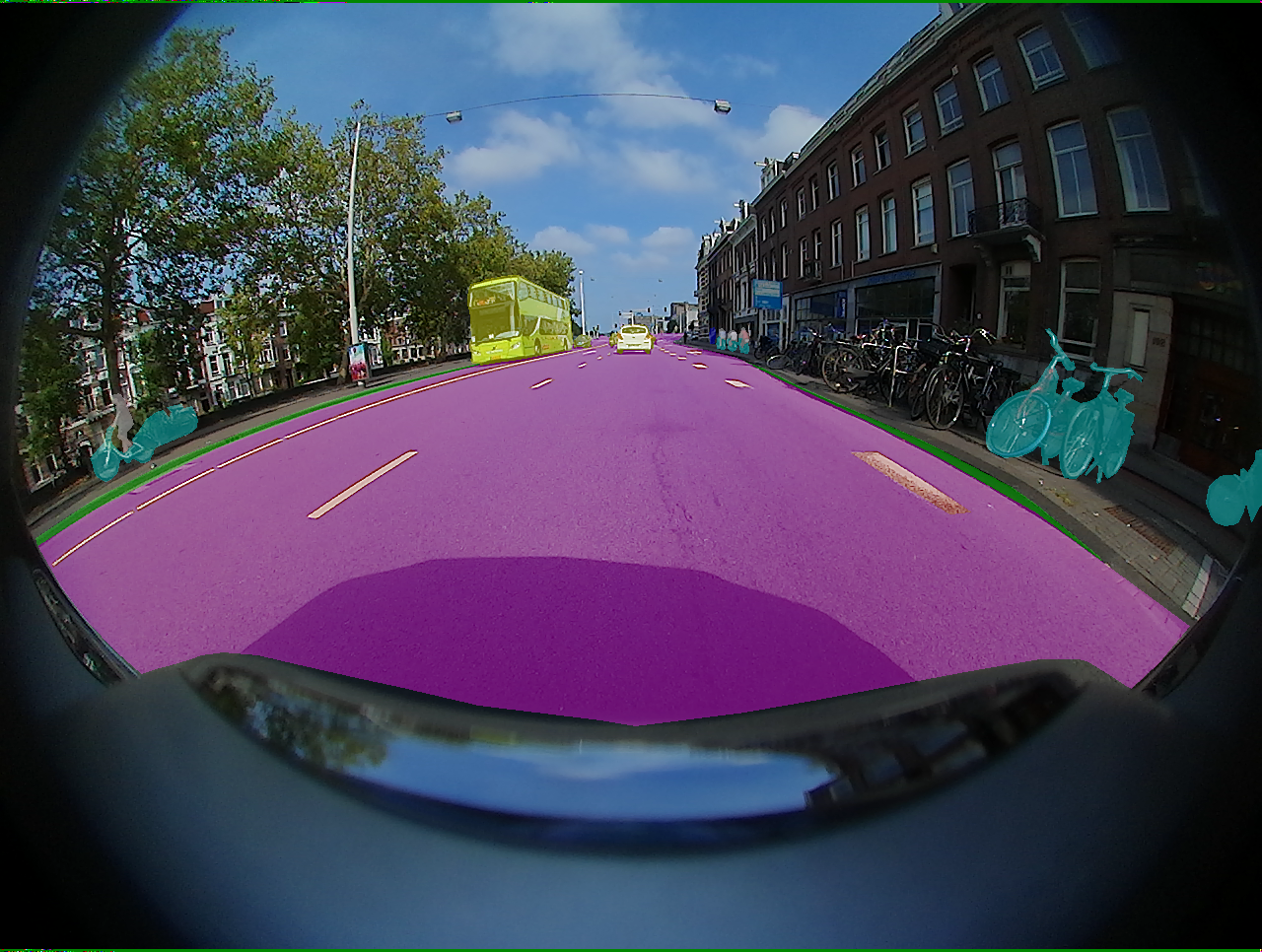}
    \caption{Some (but not all) parked bicycles are labeled as \emph{bicycle}.}
    \vspace{\baselineskip}
  \end{subfigure}
  \begin{subfigure}{\textwidth}
    \centering
    \includegraphics[width=0.35\textwidth]{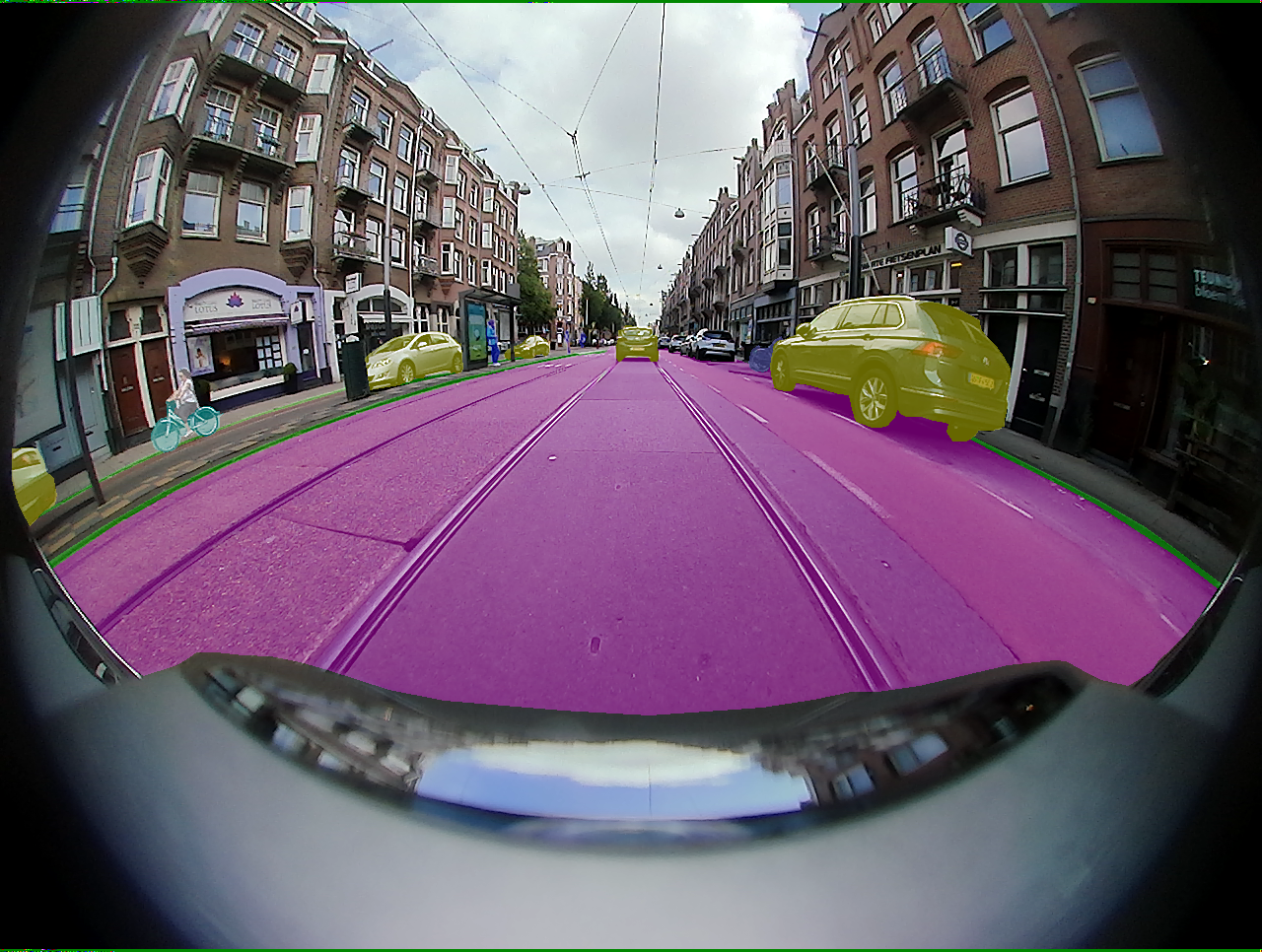}
    \hspace{1em}
    \includegraphics[width=0.35\textwidth]{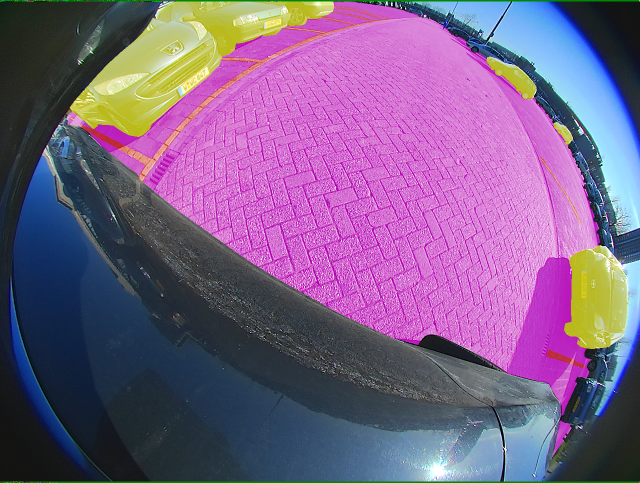}
    \caption{Some (but not all) parked cars are labeled as \emph{vehicles}.}
  \end{subfigure}
  \caption{Examples of inconsistencies in semantic masks of the WoodScape dataset.}
  \label{fig:ws_problems}
\end{figure*}

\paragraph{Stanford 2D-3D-S} The Stanford 2D-3D-Semantic dataset\footnote{\url{http://buildingparser.stanford.edu/dataset.html}}~\cite{Armeni2017Stanford} consists of 1413 omni-directional RGB-D images of indoor scenes from six different buildings. These six areas are used to create an official three fold cross validation split. Each area has complete semantic segmentation annotations for 13 object classes and 2 "void" classes: "background" and "unknown". For all tasks we map all regions of the "unknown" class to the "background" class.  We normalize all RGB-D input channels individually and ignore the background class during training and evaluation, in line with~\cite{Jiang20219SphericalCNNUnstructuredGrids}.

\begin{figure*}
\centering
\begin{subfigure}{0.48\textwidth}
    \adjustbox{margin=0mm 20mm 0mm 0mm}{
        \includegraphics[width=0.95\textwidth]{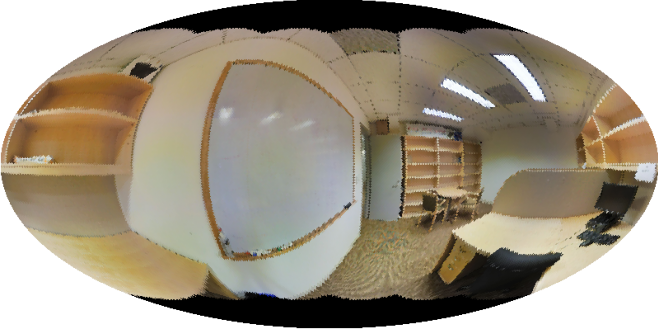}
    
    }
    \caption{The RGB channels. For the polar regions all RGB values are zero and hence here shown in black.}
    \label{fig:stanford_rgb_sample}
\end{subfigure}
\hfill
\begin{subfigure}{0.48\textwidth}
    \includegraphics[width=\textwidth]{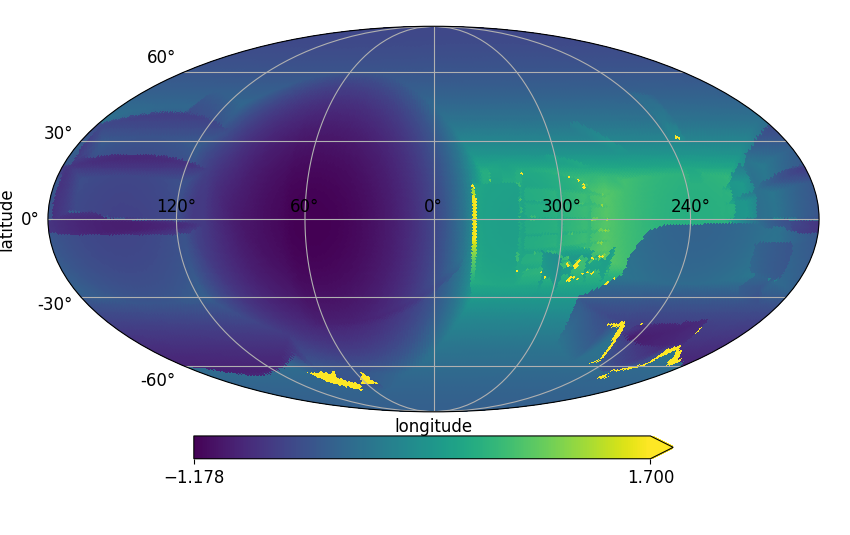}
    \caption{Log of the depth channel. Unknown depth information is represented as values above $\sim65$m, in the figure shown as solid yellow. Those areas are also prescribed the unknown class in the ground truth (mapped to background during training) which are shown in dark blue in Subfigure (c). Note that the polar regions have valid depth values.}
    \label{fig:stanford_depth_sample}
\end{subfigure}\\
\begin{subfigure}{0.48\textwidth}
    \includegraphics[width=\textwidth]{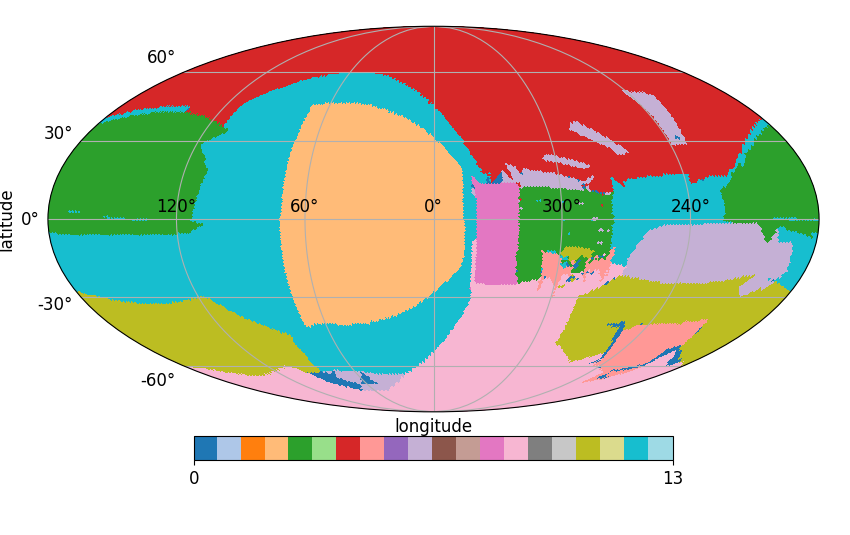}
    \caption{Full semantic ground truth. Note that the polar regions have full semantic ground truth even though the RGB channels lack information in these areas.}
    \label{fig:polar_gt_kept}
\end{subfigure}
\hfill
\begin{subfigure}{0.48\textwidth}
    \adjustbox{margin=0mm 4mm 0mm 0mm}{
        \includegraphics[width=\textwidth]{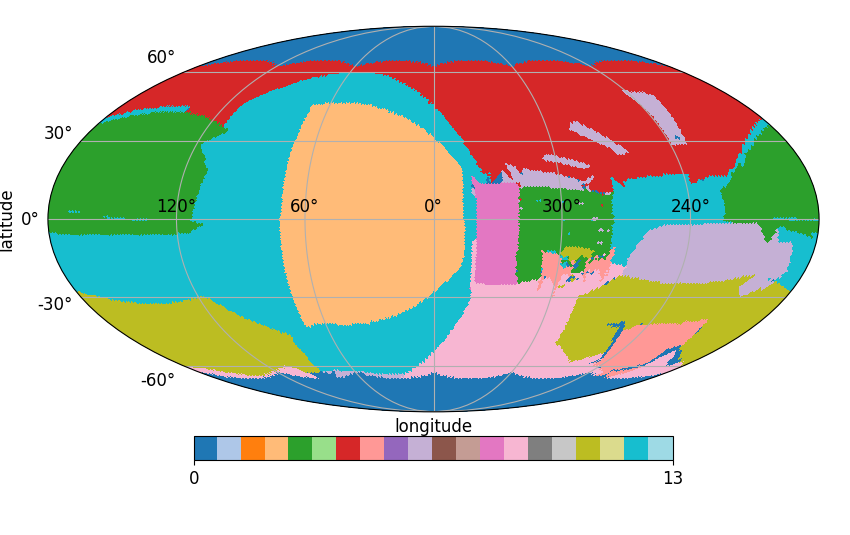}
    }
    \caption{Semantic ground truth with the polar regions mapped to background. This is what the model is trained on.}
    \label{fig:polar_gt_removed}
\end{subfigure}
\caption{Visualisation of a RGBD sample from the Stanford 2D-3D-S dataset. Subfigure (a) shows the RGB channels while Subfigure (b) displays the depth channels, where areas with unknown depth are shown in solid yellow. These together form the input to the network. Subfigure (c) shows the full semantic segmentation ground truth with ground truth also in the polar regions. Note that the areas corresponding to background/the unknown class, shown in dark blue, are the same areas that have unknown depth in Subfigure (b). Subfigure (d) shows the semantic ground truth after the polar regions have been mapped to background which is what the model is trained on.
}    
\label{fig:stanford_sample}
\end{figure*}

In contrast to the depth values and the semantic segmentation ground truth, which exist for the entire sphere, the image RGB values are non-zero only between $-60^\circ$ and $60^\circ$, see Figure~\ref{fig:stanford_sample} for a visualization. For training we chose to map the polar regions where no non-zero RGB data is present to background which is ignored during training. In that way the only areas the network is trained on are those where both RGB and depth values are present.

Removing the polar ground truth affects the computation of the IoU; specifically for classes that are heavily correlated with these areas. 
In the case of the Stanford 2D-3D-S dataset both the ceiling and floor classes often have large overlap with the polar regions and hence those classes have a disproportionately large union, which, in turn, will reduce their IoU-score.

For completeness, we report the IoU for the same trained instance of the mode on both cases: when the polar regions are mapped to background and when the segmantic ground truth is kept. The per class IoU metrics for both cases are presented in Table~\ref{tab:stanford_dataset_iou_per_class}.

In addition, there are degenerate samples in the Stanford 2D-3D-S dataset in  which the ground truth consists only of the background class. One such sample is shown in Figure~\ref{fig:worst_stanford_pred}.

\begin{table*}
  \setlength{\tabcolsep}{4pt}
  \centering
  
  \caption{Per-class intersection over union of spherical models on the Stanford 2D-3D dataset. The same instance of HEAL-SWIN is evaluated for the cases where the polar ground truth is kept ($\dagger$) and where it is mapped to background ($*$).}

  \label{tab:stanford_dataset_iou_per_class}
  \footnotesize
\begin{tabular}{c|c|ccccccccccccccc}
\toprule
Method                                                & mIoU          & beam          & board         & bookcase      & ceiling       & chair         & clutter       & column        & door          & floor         & sofa          & table         & wall          & window \\
\midrule
Gauge CNN~\cite{cohen2019gauge}                       & 39.4          & –             & –             & –             & –             & –             & –             & –             & –             & –             & –             & –             & –             & –\\
UGSCNN~\cite{Jiang20219SphericalCNNUnstructuredGrids} & 38.3          & 8.7           & 32.7          & 33.4          & 82.2          & 42.0          & 25.6          & 10.1          & 41.6          & 87.0          & 7.6           & 41.7          & 61.7          & 23.5\\
HexRUNet~\cite{zhang2019b}                            & 43.3          & 10.9          & 39.7          & 37.2          & \textbf{84.8} & 50.5          & 29.2          & 11.5          & 45.3          & \textbf{92.9} & 19.1          & 49.1          & 63.8          & 29.4\\
SphCNN~\cite{esteves2018, esteves2020c}               & 40.2          & -             & -             & -             & -             & -             & -             & -             & -             & -             & -             & -             & -             & -  \\
Spin-SphCNN~\cite{esteves2020c}                       & 41.9          & -             & -             & -             & -             & -             & -             & -             & -             & -             & -             & -             & -             & -  \\
\midrule
HEAL-SWIN$^*$                                             & 44.3          & \textbf{11.8} & \textbf{42.8} & \textbf{42.0} & 67.2          & 57.8          & \textbf{33.9} & \textbf{12.9} & 50.9          & 66.0          & \textbf{24.5} & \textbf{56.8} & 68.7          & \textbf{40.2} \\
HEAL-SWIN$^\dagger$                                   & \textbf{47.3} & 11.5          & 42.8          & 42.0          & 83.8          & \textbf{58.8} & 33.8          & 12.8          & \textbf{52.0} & 87.6          & 24.4          & 56.8          & \textbf{68.8} & 40.2\\
\bottomrule
\end{tabular}
\end{table*}

\begin{table*}
  \setlength{\tabcolsep}{4pt}
  \centering
  \caption{Per-class accuracy of spherical models on the Stanford 2D-3D dataset.}
  \label{tab:stanford_dataset_acc_per_class}
  \footnotesize
\begin{tabular}{c|c|ccccccccccccccc}
\toprule
Method                                                & mAcc          & beam          & board         & bookcase      & ceiling       & chair         & clutter       & column        & door          & floor         & sofa          & table         & wall          & window \\
\midrule
Gauge CNN~\cite{cohen2019gauge}                       & 55.9          & –             & –             & –             & –             & –             & –             & –             & –             & –             & –             & –             & –             & –\\
UGSCNN~\cite{Jiang20219SphericalCNNUnstructuredGrids} & 54.7          & 19.6          & 48.6          & 49.6          & 93.6          & 63.8          & 43.1          & \textbf{28.0} & 63.2          & 96.4          & 21.0          & 70.0          & 74.6          & 39.0\\
HexRUNet~\cite{zhang2019b}                            & 58.6          & \textbf{23.2} & 56.5          & \textbf{62.1} & 94.6          & 66.7          & 41.5          & 18.3          & 64.5          & 96.2          & 41.1          & \textbf{79.7} & 77.2          & 41.1\\
SphCNN~\cite{esteves2018, esteves2020c}               & 52.8          & -             & -             & -             & -             & -             & -             & -             & -             & -             & -             & -             & -             & -  \\
Spin-SphCNN~\cite{esteves2020c}                       & 55.6          & -             & -             & -             & -             & -             & -             & -             & -             & -             & -             & -             & -             & -  \\
\midrule
HEAL-SWIN                                             & \textbf{61.9} & 18.9          & \textbf{58.3} & 61.0          & \textbf{95.6} & \textbf{75.4} & \textbf{50.9} & 20.2          & \textbf{66.5} & \textbf{97.7} & \textbf{41.3} & 76.7          & \textbf{88.9} & \textbf{52.7} \\
\bottomrule
\end{tabular}
\end{table*}

\newlength{\subfigwidth}
\setlength{\subfigwidth}{0.45\textwidth}

\begin{figure*}
    \centering
    \begin{subfigure}{\subfigwidth}
        \includegraphics[width=\textwidth]{graphics/stanford_visualisations/cross_val_fold_1/keep_polar_gt/best_0_iou_bg_removed_0.812_cross_val_fold-1_polar_cap_removed_False_n5c_office_35_img.png}
        \caption{The RGB channels.}
    \end{subfigure}\\
    \begin{subfigure}{\subfigwidth}
        \includegraphics[width=\textwidth]{graphics/stanford_visualisations/cross_val_fold_1/keep_polar_gt/best_0_iou_bg_removed_0.812_cross_val_fold-1_polar_cap_removed_False_n5c_office_35_gt.png}
        \caption{The semantic ground truth.}
    \end{subfigure}\hfill
    \begin{subfigure}{\subfigwidth}
        \includegraphics[width=\textwidth]{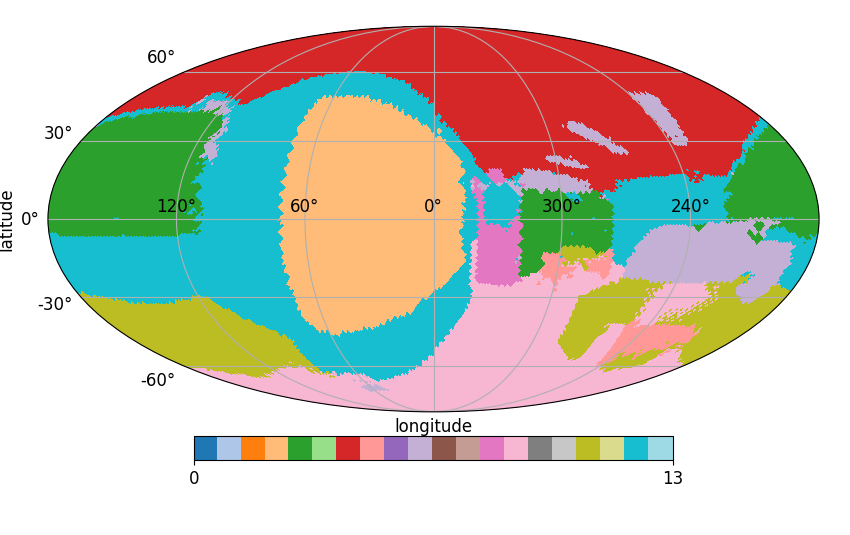}
        \caption{The predicted segmentation.}
    \end{subfigure}
    \caption{One of the best per sample predictions. Recall that during training the model did not update on areas lacking RGB data. This sample is from the evaluation set in cross validation fold 1.}
    \label{fig:best_stanford_pred}
\end{figure*}

\begin{figure*}
    \centering
    \begin{subfigure}{\subfigwidth}
        \includegraphics[width=\textwidth]{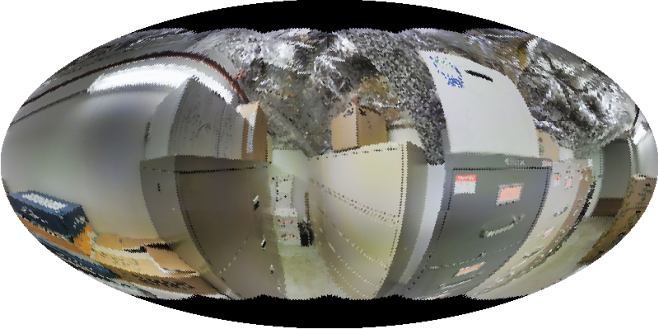}
        \caption{The RGB channels. Note that this sample is very visually noisy.}
    \end{subfigure}\\
    \begin{subfigure}{\subfigwidth}
        \includegraphics[width=\textwidth]{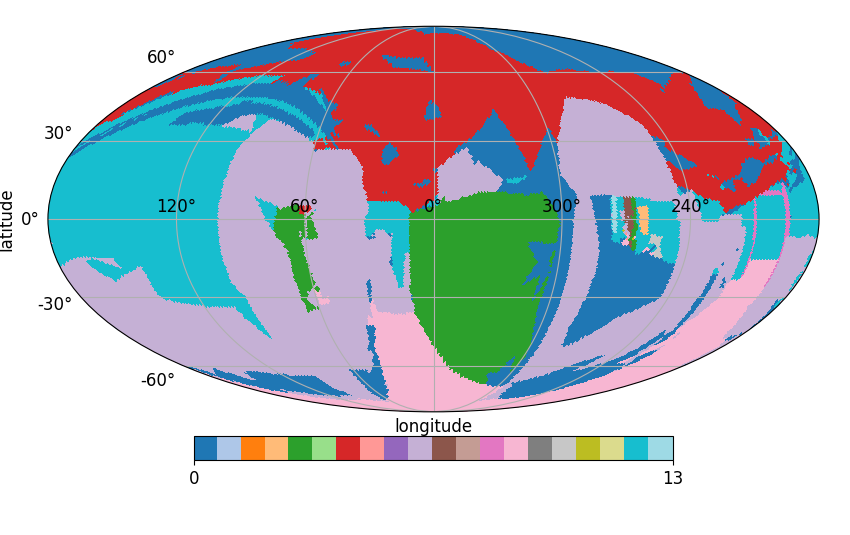}
        \caption{The semantic ground truth. Note that large irregular areas are unknown (and hence mapped to background).}
    \end{subfigure}\hfill
    \begin{subfigure}{\subfigwidth}
        \includegraphics[width=\textwidth]{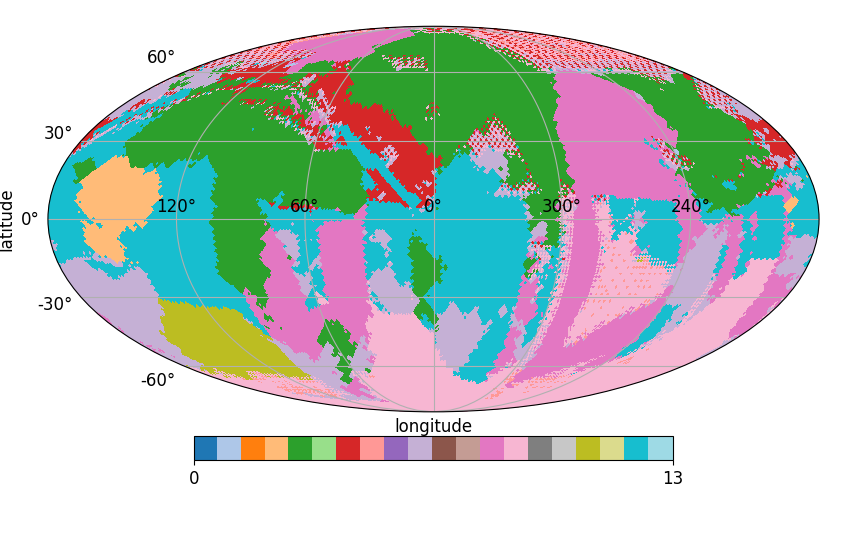}
        \caption{The predicted segmentation.}
    \end{subfigure}
    \caption{One of the worst per sample predictions. Recall that during training the model did not update on areas lacking RGB data. This sample is from the evaluation set in cross validation fold 1.}
    \label{fig:worst_legitimate_stanford_pred}
\end{figure*}

\begin{figure*}
    \centering
    \begin{subfigure}{\subfigwidth}
        \includegraphics[width=\textwidth]{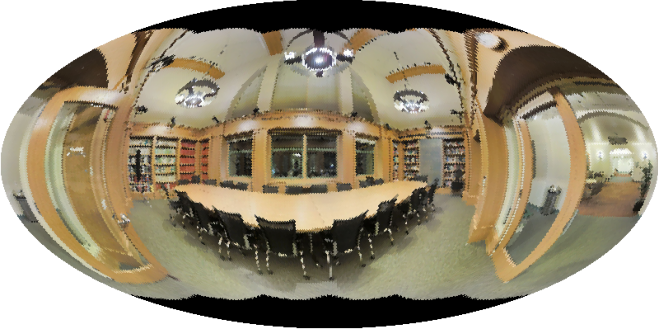}
        
        \caption{RGB channels.}
    \end{subfigure}\\
    \begin{subfigure}{\subfigwidth}
        \includegraphics[width=\textwidth]{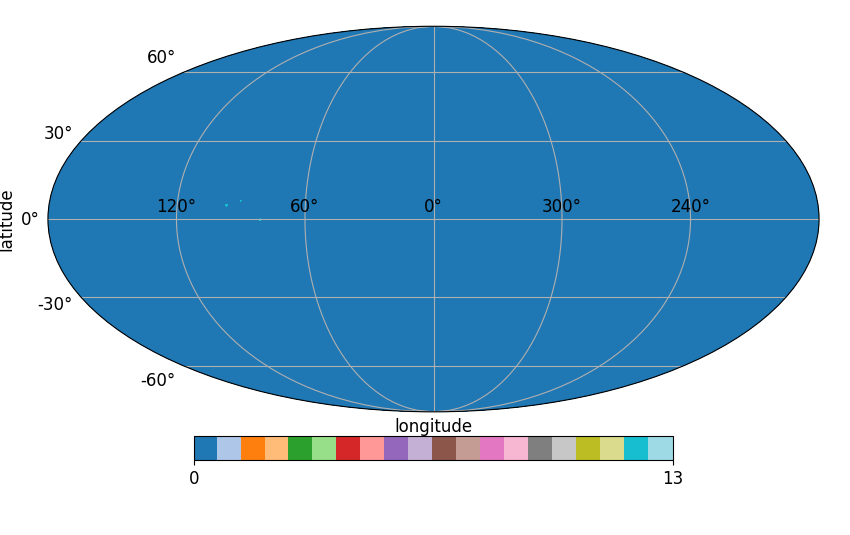}
        \caption{Ground truth.}
    \end{subfigure}\hfill
    \begin{subfigure}{\subfigwidth}
        \includegraphics[width=\textwidth]{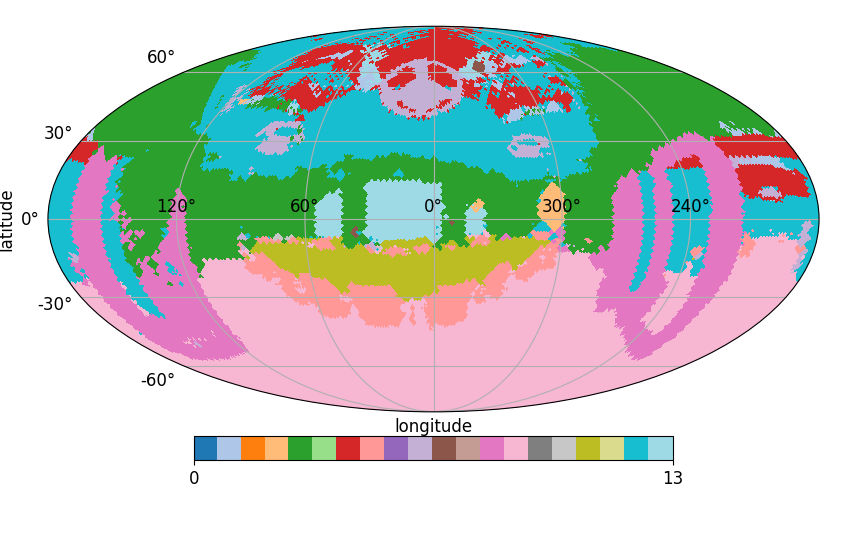}
        \caption{Predicted segmentation.}
    \end{subfigure}
    \caption{The worst per sample prediction. Note that the predicted segmentation is reasonable but that the ground truth, for some reason, consists of only the background class. This sample is from the evaluation set in cross validation fold 3. Although neither this nor similar samples were removed from the dataset during training, they were effectively ignored since the background class is ignored during training.}
    \label{fig:worst_stanford_pred}
\end{figure*}

\section{Additional experiment: Spherical image classification}
\label{app:additional_experiments}

\begin{table*} 
  \centering
  \caption{Classification accuracy on spherical MNIST when trained and evaluated on non-rotated data (NR/NR) and on rotated data (R/R). Equivariant models are marked with an asterisk.}
  \label{tab:s2_mnist}
  \begin{tabular}{ccc}
    \toprule
    Model & NR/NR Acc & R/R Acc\\
    \midrule
    S2CNN*~\cite{cohen2018b} & 96 & 95 \\
    Clebsch-Gordan Nets*~\cite{kondor2018} & 96.4 & 96.6 \\
    Gauge CNN*~\cite{cohen2019gauge} & 99.43 & {\bf 99.31} \\    
    SphCNN*~\cite{esteves2018, esteves2020c} & 98.75 & 98.71 \\
    Spherical Transformer*~\cite{Cho2022SphericalTransformer} & -- & 95.09 \\
    \midrule
    UGSCNN~\cite{Jiang20219SphericalCNNUnstructuredGrids} & 99.23 & 94.92 \\
    HexRUnet~\cite{zhang2019b} & {\bf 99.45} & 97.05 \\
    HEAL-SWIN (Ours) & 99.20 & 96.96 \\ 
    \bottomrule
  \end{tabular}
\end{table*}
A common low-resolution dataset on which performance of spherical models is measured is a spherical projection of MNIST. We project the MNIST digits onto a HEALPix grid of $\nside=16$, corresponding to 3072 input pixels, less than the 3600 input pixels often used on the Driscoll-Healy grid. On this dataset, we train a HEAL-SWIN classifier consisting of 10 transformer layers followed by three fully-connected layers resulting in a model with about 62k parameters.

We train and evaluate our model both on unrotated data (NR/NR modality) and on rotated data (R/R modality). In the NR/NR modality, the task is very simple and most spherical models (including ours) reach nearly perfect performance, as shown in Table~\ref{tab:s2_mnist}. The R/R modality, in which the images are rotated by a random rotation in $\mathrm{SO}(3)$, is specifically designed for testing equivariant models. Therefore, these have a substantial advantage since they do not need to learn the symmetry of the task. In the R/R modality, our model is only outperformed by some equivariant models and performs better than or on par with all other models. Note, however, that the equivariant models do not scale to high-resolution inputs.

\section{Experimental details}
\label{app:heal-swin-model-details}
\subsection{SynWoodscape and MNIST experiments}
In Table~\ref{tab:heal_swin_spatial_features} we provide further details on the spatial size of the features throughout the HEAL-SWIN model used in the experiments discussed in Section~\ref{sec:experiments}.

\paragraph{Resolution} In order to eliminate resolution as a central parameter in comparing the HEAL-SWIN to the SWIN, we first rescale the input images to a size of $640\times 768$ giving a resolution of approximately $492$k, which we can sample to the HEALPix grid using $\nside=256$ yielding a resolution of around $\sim 525$k.

\begin{table*}
    \centering
    \caption{Spatial features per layer in the HEAL-SWIN models used for the experiments described in Section~\ref{sec:experiments}.}
    \label{tab:heal_swin_spatial_features}
    \vspace{\baselineskip}
    \begin{tabular}{cccccc}
        \toprule
        layer & pixel / patches & windows & \makecell{windows per\\base pixel} & $\nside$ & followed by \\
        \midrule
        input & 524288 & 8192 & 1024 & 256 & patch embedding \\
        HEAL-SWIN block 1 & 131072 & 2048 & 256 & 128 & patch merging \\
        HEAL-SWIN block 2 & 32768 & 512 & 64 & 64 & patch merging \\
        HEAL-SWIN block 3 & 8192 & 128 & 16 & 32 & patch merging \\
        HEAL-SWIN block 4 & 2048 & 32 & 4 & 16 & patch expansion \\
        HEAL-SWIN block 5 & 8192 & 128 & 16 & 32 & patch expansion \\
        HEAL-SWIN block 6 & 32768 & 512 & 64 & 64 & patch expansion \\
        HEAL-SWIN block 7 & 131072 & 2048 & 256 & 128 & patch expansion \\
        output & 524288 & 8192 & 1024 & 256 & --- \\
        \bottomrule
    \end{tabular}
\end{table*}

\paragraph{Hardware and training details} For the semantic segmentation task we train all models on four \texttt{Nvidia A40} GPUs with an effective batch size of 8 and a constant learning rate of $9.4 \times 10^{-4}$. For the depth estimation task we used an effective batch size of 4 and learning rates of $5 \times 10^{-3}$ and $5 \times 10^{-5}$ for the HEAL-SWIN and SWIN models, respectively, chosen from the best performing models after a learning rate ablation.

\paragraph{Classes and reweighting} We adjust the number of output channels in the base HEAL-SWIN and SWIN models described in Section~\ref{sec:experiments} to the number of classes and train with a weighted pixel-wise cross-entropy loss. We choose the class weights $w_i$ to be given in terms of the class prevalences $n_i$ by $w_i = n^{-1/4}_i$.

\paragraph{HEAL-SWIN versus SWIN for flat segmentation}
\label{app:flat_sem_seg_results}

In Table~\ref{tab:flat_segmentation}, we show the results of evaluating the segmentation models discussed in Section~\ref{subsec:sem_seg} on the plane. In this case, the HEAL-SWIN predictions are projected onto the pixel grid of the SWIN predictions before evaluation. To ensure a fair comparison, the flat mIoU is calculated on a masked region of this grid, removing pixels which lie outside of the (restricted) HEALPix grid we use.

\subsection{Stanford 2D-3D-S experiments}

\paragraph{Resolution} In order to be close to the resolution used by HexRUNet~\cite{zhang2019b}, we chose  $\nside=64$ which resulted in 49k pixels in the HEALPix grid, compared to 20k pixels used for HexRUNet. 

\paragraph{Hardware and training details} 
Training was conducted with a constant learning rate of $5\times 10^{-3}$ on four \texttt{Nvidia A100} due to the smaller demand on them compared to the \texttt{Nvidia A40} on the compute cluster, although the \texttt{A40}'s would work perfectly fine. The experiments on the Stanford 2D-3D-S dataset used an effective batch size of 80, a small weight decay of 0.1 and a gradient clipping on 0.5 acting on the total gradient 2-norm.

\paragraph{Additional results}
For a per-class performance breakdown comparing the HEAL-SWIN model to previous comparable models, see Table~\ref{tab:stanford_dataset_iou_per_class} and \ref{tab:stanford_dataset_acc_per_class}.

In Figure~\ref{fig:best_stanford_pred} and \ref{fig:worst_legitimate_stanford_pred} we show the best and worst predictions of our model respectively.

\paragraph{Model architecture}
For a good comparison to HexRUNet~\cite{zhang2019b} and UGSCNN~\cite{Jiang20219SphericalCNNUnstructuredGrids}, which have 1.5M and 5.2M parameters, respectively, we construct a HEAL-SWIN model with 1.5M parameters. 
Table~\ref{tab:stanford_heal_swin_spatial_features} shows the spatial features per block. We performed ablations to set window, patch, and shift sizes, see Table~\ref{tab:ablations}.

\begin{table*}
  \centering
  \caption{Ablations over patch size, window size, shift size and shifting strategy on the Stanford 2D-3D. Performance is measured using the three-fold cross-validation of that dataset. Unless otherwise stated, the model parameters are $\npatch=4$, $\nwindow=64$, spiral shifting with $\nshift=4$ and the model architecture is the same whose performance is reported in Table~\ref{tab:stanford_dataset}.}
  \label{tab:ablations}
  \begin{tabular}{llcc}
    \toprule
    \multicolumn{2}{c}{Parameter values} & mIoU & mAcc \\
    \midrule
    \multicolumn{2}{l}{$\npatch=4$} & 43.2 & 61.1 \\
    \multicolumn{2}{l}{$\npatch=16$} & 40.9 & 58.9 \\
    \midrule
    
    $\nwindow=64$ & $\nshift=4$ & 43.2 & 61.1 \\ 
    
    $\nwindow=16$ & $\nshift=2$ & 44.3 & 61.9\\  
    \midrule
    \multicolumn{2}{l}{$\nshift=2$} & 42.2 & 60.5 \\ 
    \multicolumn{2}{l}{$\nshift=4$} & 43.2 & 61.1 \\
    \multicolumn{2}{l}{$\nshift=8$} & 39.3 & 56.6 \\ 
    \midrule
    \multicolumn{2}{l}{spiral shifting} & 43.2 & 61.1 \\ 
    \multicolumn{2}{l}{grid shifting} & 42.3 & 59.9 \\ 
    \bottomrule
  \end{tabular}
\end{table*}

\begin{table*}
    \centering
    \caption{Spatial features per layer in the HEAL-SWIN models used for the Stanford 2D3Ds experiments.}
    \label{tab:stanford_heal_swin_spatial_features}
    \vspace{\baselineskip}
    \begin{tabular}{cccccc}
        \toprule
        layer & pixel / patches & windows & \makecell{windows per\\base pixel} & $\nside$ & followed by \\
        \midrule
        input & 49152 & 1536 & 256 & 64 & patch embedding \\
        HEAL-SWIN block 1 & 12288 & 768 & 64 & 32 & patch merging \\
        HEAL-SWIN block 2 & 3072 & 192 & 16 & 16 & patch merging \\
        HEAL-SWIN block 3 & 768 & 48 & 4 & 8 & patch expansion \\
        HEAL-SWIN block 4 & 3072 & 192 & 16 & 16 & patch expansion \\
        HEAL-SWIN block 5 & 12288 & 768 & 64 & 32 & patch expansion \\
        output & 49152 & 1536 & 256 & 64 & --- \\
        \bottomrule
    \end{tabular}
\end{table*}

\paragraph{Computational complexity}
To verify the limited computational overhead of the HEAL-SWIN architecture we provide ablations over a range of resolutions in Table \ref{tab:computational}.
\begin{table*} 
\caption{
\label{tab:computational}
Comparison of inference times for HEAL-SWIN and SWIN ablated over data resolution.
}
\centering
\begin{tabular}{rlll}
\toprule
& Resolution & Pixels & time / pixel \\
\midrule
HEAL-SWIN & $8\times 256.0^2 $& $5.2\cdot 10^5$ & 297 $\pm$ 26ns \\
SWIN      & $640\times 768   $& $4.9\cdot 10^5$ & 296 $\pm$ 39ns \\
\midrule
HEAL-SWIN & $8\times 128.0^2 $& $1.3\cdot 10^5$ & 559 $\pm$ 127ns \\
SWIN      & $256\times 384   $& $1.0\cdot 10^5$ & 660 $\pm$ 171ns \\
\midrule
HEAL-SWIN & $8\times 64.0^2  $& $0.3\cdot 10^5$ & 2031 $\pm$ 519ns \\
SWIN      & $128\times 160   $& $0.2\cdot 10^5$ & 2792 $\pm$ 668ns \\
\bottomrule
\end{tabular}
\end{table*}

\begin{table*}
  \centering
  \caption{Mean IoU for semantic segmentation with HEAL-SWIN and SWIN, averaged over three runs, after projection onto the plane. For WoodScape, we exclude the \emph{void} class from the mean but keep it in the loss.}
  \label{tab:flat_segmentation}
  \vspace{\baselineskip}  
\begin{tabular}{llc}
  \toprule
  Model & Dataset & flat mIoU \\
  \midrule
HEAL-SWIN & Large SynWoodScape & 0.899 \\ 
SWIN & Large SynWoodScape & 0.930 \\ 
\midrule
HEAL-SWIN & Large+AD SynWoodScape & 0.790 \\ 
SWIN & Large+AD SynWoodScape & 0.837 \\ 
\midrule
HEAL-SWIN & WoodScape & 0.611 \\ 
SWIN & WoodScape & 0.620 \\
\bottomrule
\end{tabular}
\end{table*}

\end{document}